\documentclass[a4paper,fleqn]{cas-dc}

\usepackage[T1]{fontenc}
\usepackage[english]{babel}
\usepackage{tgbonum}
\usepackage{amsmath,amssymb,amsthm,mathrsfs,bm}
\usepackage{graphicx}
\usepackage{booktabs}
\usepackage{multirow}
\usepackage{tabularx}
\usepackage{siunitx}
\usepackage[table]{xcolor}
\usepackage{subcaption}
\usepackage{makecell}
\usepackage{enumitem}
\usepackage[ruled,vlined]{algorithm2e}
\usepackage{orcidlink}
\usepackage{array}
\usepackage{ragged2e}

\usepackage{float}
\usepackage{placeins}
\usepackage{tikz}

\usepackage{xurl}
\newtheorem{theorem}{Theorem}

\begin{document}

\let\WriteBookmarks\relax
\def\floatpagepagefraction{1}
\def\textpagefraction{.001}

\shorttitle{SpectONet}
\shortauthors{Saini, Vats and Sahoo}

\title[mode=title]{SpectONet: A Physics-Guided Spectral Deep Operator Network for Euler-Bernoulli Beam Dynamics}

\author[a]{Shivani Saini}[
orcid=0009-0003-8679-8120
]
\credit{Conceptualization, Methodology, Software, Validation, Investigation, Writing - original draft}
\ead{shivanis.nith@gmail.com}

\author[a]{Ramesh Kumar Vats}[
orcid=0000-0002-7974-5341
]
\credit{Supervision, Methodology, Writing - review \& editing}
\ead{rkvats@nith.ac.in}

\author[b]{Arup Kumar Sahoo}[
orcid=0000-0003-4515-7434
]
\cormark[1]
\credit{Conceptualization, Methodology, Software, Validation, Investigation, Writing - original draft}

\ead{arupnitr.jrfmath@gmail.com, asahoo@campus.haifa.ac.il}

\cortext[1]{Corresponding author}

\affiliation[a]{
    organization={Department of Mathematics and Scientific Computing,
    National Institute of Technology Hamirpur},
    postcode={177005},
    city={Himachal Pradesh},
    country={India}
}

\affiliation[b]{
    organization={The Hatter Department of Marine Technologies,
    Leon H. Charney School of Marine Sciences,
    University of Haifa},
    city={Haifa},
    postcode={3498838},
    country={Israel}
}

\begin{abstract}\justifying
\noindent This paper proposes a novel physics-guided spectral deep operator network, termed SpectONet, for solving Euler-Bernoulli beam (EBB) vibration problems. The proposed framework integrates the operator-learning capability of DeepONet with physics-informed constraints and Chebyshev-Gauss-Lobatto (CGL) sensor placement. Unlike conventional DeepONet frameworks, which commonly employ uniformly distributed sensors, SpectONet uses nonuniform spectral sensor locations with a higher concentration of points near the domain boundaries. This sampling strategy improves the finite-dimensional representation of boundary-sensitive structural responses while requiring only a limited number of branch-network inputs. The governing beam equation, together with the associated initial and boundary conditions, is incorporated into the training objective to promote physically consistent and generalizable predictions. Numerical experiments on three EBB vibration problems demonstrate the effectiveness of the proposed framework. Comparisons with strong baseline models such as Vanilla DeepONet, PI-DeepONet, PINN, and CNN-UNet show that SpectONet consistently achieves lower prediction errors across all considered evaluation metrics. In particular, SpectONet achieves an average reduction of approximately $79\%$ in the mean relative $\mathcal{L}_2$ error compared with the best-performing baseline across the three benchmark problems. These results demonstrate the effectiveness of SpectONet as an accurate physics-guided operator-learning framework for structural vibration analysis.

\begin{center}
    \includegraphics[
        width=0.95\textwidth,
        height=0.34\textheight
    ]{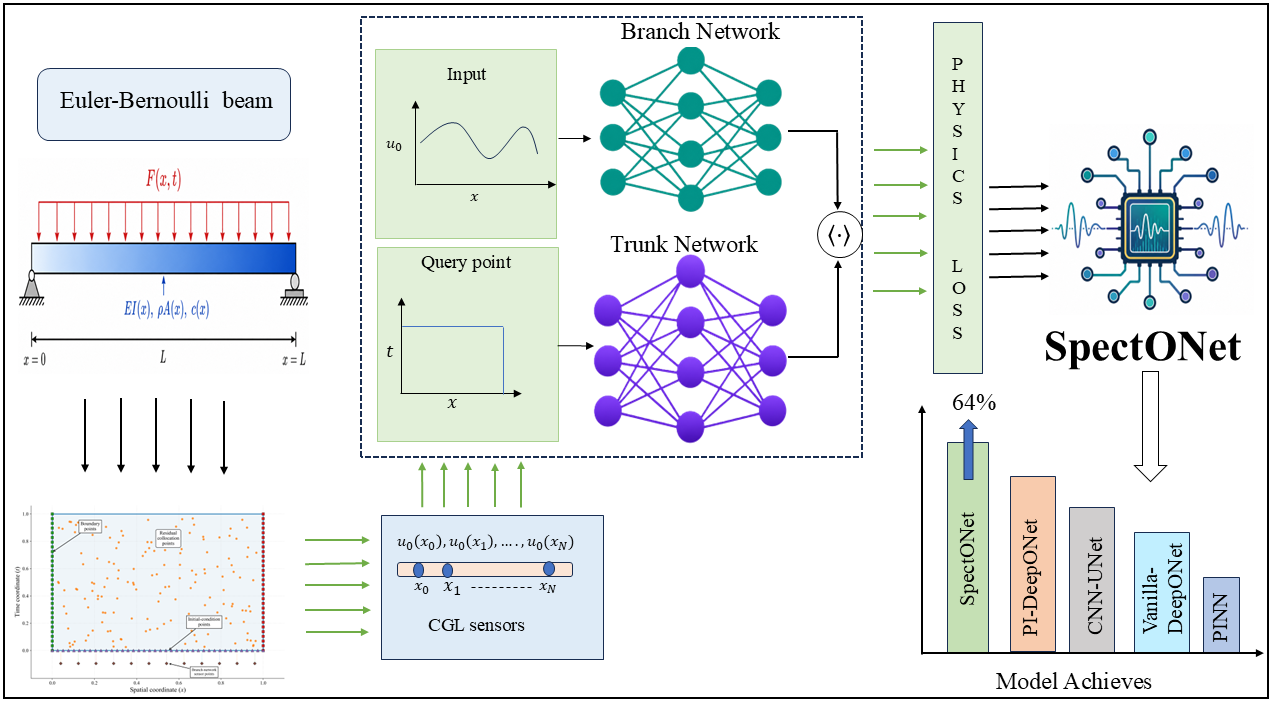}

    \small\textbf{Graphical abstract}
\end{center}
\end{abstract}

\begin{keywords}
Neural Operator \sep Scientific Machine Learning  \sep  Physics-Guided DeepONet  \sep Euler-Bernoulli Beam  \sep Structural Vibration \sep Chebyshev-Gauss-Lobatto Sensor
\end{keywords}

\maketitle

\section{Introduction}
\justifying

Artificial intelligence (AI) has emerged as one of the fastest-growing computational technologies in the $21^{st}$ century, with successful applications in pattern recognition, natural language processing, computer vision, robotics, and scientific computing~\cite{goodfellow2016deep,chakraverty2025artificial, sahoo2022machine}. Among the various AI approaches, artificial neural networks (ANNs) have gained considerable attention because of their ability to learn complex nonlinear relationships directly from available data.
Owing to their strong approximation and generalization capabilities, ANNs have been increasingly employed for modeling dynamical problems governed by differential, integral, and polynomial equations~\cite{oparaji2017robust,saini2026pinn,kumar2025deep}.

A wide variety of real-world phenomena arising in structural mechanics, fluid dynamics, heat transfer, electrostatics, wave propagation, and mathematical physics can be described using ordinary differential equations (ODEs) and partial differential equations (PDEs). In structural mechanics, such equations are extensively employed to model the static and dynamic behavior of advanced engineering structures, including multidirectional functionally graded nanostructures~\cite{gartia2025advanced}, composite beams~\cite{baysal2024exponential}, and functionally graded plates and shells~\cite{chakraverty2008vibration}. The increasing complexity of these systems, together with variations in material properties, external loading, geometrical configurations, initial conditions (ICs), and boundary conditions (BCs), motivates the development of efficient computational frameworks. In this context, ANN-based approaches provide a promising means of learning the underlying solution behavior and establishing nonlinear mappings between physical inputs and the corresponding system responses.

One of the earliest applications of ANNs to solve dynamical problems was presented by Lagaris et al.~\cite{lagaris1998artificial}. They developed a feedforward neural network framework for approximating the solutions of ODEs and PDEs while satisfying the prescribed ICs and BCs. Their work demonstrated the potential of ANNs as flexible and mesh-free computational tools for solving differential equations problems. Since then, ANNs have been widely employed as surrogate models for complex nonlinear systems because they can provide rapid predictions after training. However, conventional ANNs are primarily data-driven, typically require large labeled datasets, and may produce physically inconsistent solutions when the physical laws are not incorporated during model training.

To address these limitations, physics-informed neural networks (PINNs)~\cite{raissi2019physics} have emerged as a powerful mesh-free framework for solving forward and inverse problems. Unlike conventional data-driven neural networks, PINNs incorporate the differential equations together with the associated ICs and BCs directly into the loss function. The required spatial and temporal derivatives are generally evaluated through automatic differentiation (AD), allowing the neural network to be trained without conventional mesh-based discretization~\cite{singh2026modified, kumar2026robust}. By embedding physical knowledge into the learning process, PINNs reduce their dependence on labeled data and encourage the predicted solutions to remain consistent with the underlying governing laws.

Despite their significant advantages, standard solution-specific PINNs generally approximate a single solution corresponding to a DE configuration\cite{singh2025non}. Therefore, they typically require retraining when the governing parameters, forcing functions, IC, or BC change. Although several extended PINN frameworks have been developed to improve generalization across related problem settings, conventional PINNs do not explicitly learn the underlying solution operator. This limitation reduces their suitability for repeated-query and parameter-dependent problems.

Motivated by the need to learn entire families of solutions rather than individual solution instances, operator learning has recently emerged as a powerful paradigm for approximating mappings between infinite-dimensional function spaces. Unlike conventional learning approaches, which approximate the solution corresponding to a fixed set of governing parameters, ICs, BCs, and forcing functions, operator learning learns the mapping from input functions to the corresponding solution fields. Consequently, a trained neural operator can predict solutions for
previously unseen input functions or parameter configurations within
the learned problem family without requiring retraining for each
individual instance. Such frameworks are particularly attractive for many-query and real-time applications, where repeated evaluations of the solution operator are required. Owing to their remarkable approximation capabilities, neural operators have attracted considerable attention in computational mechanics~\cite{song2026pmno}, fluid dynamics~\cite{iglesias2026deeponet}, structural dynamics~\cite{rabinovich2026mixed}, and structural health monitoring~\cite{liu2026neural}. These capabilities make neural operators particularly suitable for complex engineering systems that require rapid prediction, repeated evaluations, and real-time decision-making.

Among the various neural operator architectures developed in recent years, the deep operator network (DeepONet), proposed by Lu et al.~\cite{lu2021learning} and theoretically supported by the universal approximation theorem for nonlinear operators, has emerged as one of the most versatile frameworks. DeepONet consists of two subnetworks: a branch network that encodes the input function and a trunk network that represents the spatial and temporal coordinates. Through the interaction of these two subnetworks, DeepONet is capable of learning nonlinear operators associated with complex PDE systems. Several neural operator architectures and extensions have subsequently been developed, including the Fourier neural operator~\cite{li2020fourier},
wavelet neural operator~\cite{tripura2023wavelet}, Laplace neural
operator~\cite{cao2024laplace}, and convolution neural
operator~\cite{liu2026physics}, each possessing distinct
advantages depending on the application domain~\cite{lone2026alpha}.

Despite their success, conventional supervised DeepONet models generally rely on paired
input-output training data generated from expensive numerical simulations or experimental measurements. The acquisition of such datasets becomes increasingly challenging and costly for complex engineering systems. To alleviate this dependency on labeled data, Wang et al.~\cite{wang2021learning} introduced the physics-informed deep operator network (PI-DeepONet), which embeds the governing equations and the associated ICs and BCs into the loss function. By incorporating physical constraints into the learning process, PI-DeepONet can achieve accurate predictions with significantly fewer training samples.

Nevertheless, the following gaps are present in existing PI-DeepONet frameworks
\begin{enumerate}
\item The accuracy of DeepONet strongly depends on the placement of sensor points used to discretize the input functions. Uniformly distributed sensors often fail to capture localized variations effectively, leading to reduced approximation accuracy and poor generalization. 
\item Although the branch network jointly processes the complete vector of sensor values, conventional PI-DeepONet frameworks commonly employ uniformly spaced sensors and therefore do not exploit the interpolation advantages of nonuniform Chebyshev-Gauss-Lobatto (CGL) node distributions, particularly their increased resolution near the domain boundaries. 
\item Equidistant sensor locations generally require a large number of sampling points to accurately represent smooth solutions and forcing functions, thereby increasing the computational cost and training complexity.
\end{enumerate}

In this regard, spectral approximation techniques based on orthogonal polynomials provide an alternative for function representation~\cite{trefethen2000spectral}. Among these techniques, CGL points exhibit excellent interpolation properties, reducing interpolation errors near the domain boundaries while providing enhanced boundary resolution. Owing to these characteristics, CGL points have been widely employed in spectral methods because they offer high approximation accuracy with relatively few collocation points~\cite{heydari2022chebyshev}. Despite their success in numerical analysis and spectral methods~\cite{kopriva2009implementing}, the incorporation of CGL-based sensor placement into physics-informed operator learning frameworks has received limited attention. This observation motivates the development of a more effective operator learning framework that leverages the advantages of spectral sensor distributions while preserving the physical consistency provided by PI-DeepONet.

To fill these gaps, this work proposes a novel physics-guided spectral deep operator network (SpectONet) for solving Euler-Bernoulli Beam (EBB) vibration problems. The proposed framework combines the operator-learning capability of DeepONet with the physical constraints imposed by PI-DeepONet and employs CGL sensor locations to improve the discretization of input functions. The branch network receives the values of the forcing function or IC sampled at CGL points, whereas the trunk network encodes the spatio-temporal coordinates. 
Unlike conventional DeepONet models and baselines that rely on supervised solution data, the proposed algorithm is trained solely using the governing physics, BC, and IC, without requiring paired input-output solution fields. Despite the absence of supervised solution loss, the proposed framework achieves superior predictive accuracy through physics-constrained operator learning and CGL-distributed sensors.

The major contributions of this research are summarized as follows:

\begin{enumerate}

\item A novel SpectONet is proposed by integrating PI-DeepONet with CGL sensor placement for solving EBB vibration problems.

\item Non-uniform CGL sensors locations are incorporated into the branch network to improve the representation of input functions and prediction accuracy.

\item  The proposed model eliminates the need for supervised solution data during training. Instead, learning is driven entirely by the governing PDE together with the associated IC and BC.

\item Extensive numerical experiments on three benchmark
EBB vibration problems demonstrate that SpectONet
consistently outperforms the baseline models in terms
of prediction accuracy, achieving an improvement of
up to 79\% in the relative $\mathcal{L}_2$ error.

\item An open-source code \url{https://github.com/SM-a-RT-Lab/SpectONet} to promote the future experiment and extension.

\end{enumerate}

The remainder of this paper is organized as follows. Section~\ref{sec_2} reviews the related work. Section~\ref{sec_3} presents the preliminaries, including the mathematical formulation of the EBB vibration problems and the fundamentals of PINN and DeepONet. Section~\ref{sec_4} describes the proposed SpectONet framework. Section~\ref{sec_5} presents the numerical experiments, followed by a detailed discussion in Section~\ref{sec_6}. Finally, Section~\ref{sec_8} concludes the paper and outlines future research directions. For interested readers, the training algorithms of the baseline models considered in this study are provided in Appendix~\ref{Apndx}.

\section{Related Work}\label{sec_2}\justifying

This section reviews prior research on DeepONet, with particular emphasis on recent methodological advancements in PI-DeepONet and its applications to structural dynamics and vibration analysis.

In a pioneering work, Goswami et al.~\cite{goswami2023physics} highlighted the potential of physics-informed operator learning by integrating governing physical laws into neural operator architectures, thereby improving predictive accuracy and generalization for forward and inverse PDE problems. Building upon this foundation, several studies have proposed enhanced neural operator frameworks to improve robustness, scalability, computational efficiency, and predictive performance. Mandl et al.~\cite{mandl2025separable} proposed separable PI-DeepONet (Sep-PI-DeepONet), a computationally efficient extension of PI-DeepONet for solving high-dimensional PDEs. The framework employed separable network architectures and forward-mode AD to alleviate the curse of dimensionality and significantly reduce computational costs compared with the PI-DeepONet framework. Similarly, Michałowska et al.~\cite{michalowska2024neural} combined DeepONet with recurrent neural networks and a moving-window strategy to improve long-term prediction of dynamical systems. Yang et al.~\cite{yang2026dd} introduced DD-DeepONet, a domain decomposition framework that learns subdomain-wise operators for efficient PDE solutions on complex geometries. The proposed method reduced training difficulty and computational cost while enabling rapid prediction under varying geometrical and physical conditions. Karumuri et al.~\cite{karumuri2026physics} developed a physics-informed latent neural operator (PI-Latent-NO) that integrates latent-space learning with physics-based constraints for time-dependent parametric PDEs. The framework employed coupled DeepONet architectures to achieve accurate, scalable, and data-efficient predictions while significantly reducing computational costs. More recently, Sarthak and Srinivasacharya~\cite{sarthak2026physics} proposed a PI-DeepONet framework for long-time simulation of nonlinear PDEs using temporal domain decomposition. By recursively applying the learned operator over successive time subdomains and employing Gaussian random field sampling, the proposed approach improved long-time prediction stability and accelerated convergence. For a broader perspective, Table~\ref{tab:operator_applications} summarizes representative neural operator frameworks reported in the literature, together with their application domains and key methodological features.

\begin{table*}
\centering
\caption{Representative neural operator frameworks, their application domains, and key methodological features.}
\label{tab:operator_applications}
\renewcommand{\arraystretch}{1.4}

\begin{tabular}{
p{0.18\textwidth}
p{0.25\textwidth}
p{0.50\textwidth}
}
\hline
\textbf{Variant} &
\textbf{Application Area} &
\textbf{Features} \\
\hline

VS-WNO~\cite{garg2024neuroscience} &
Computational mechanics &
Integrates variable spiking neurons into a wavelet neural operator
to reduce computational energy while maintaining accurate PDE
operator learning. \\
\hline
PI-DeepONet~\cite{cong2026respecting} &
Time-dependent PDEs &
Introduces a causality-aware segmental training strategy by decomposing
the temporal domain and adaptively weighting the residual losses to
improve convergence and predictive accuracy. \\
\hline

Phase-Field DeepONet~\cite{li2023phase} &
Phase-field dynamics &
Combines PI-DeepONet, minimizing movement schemes, and energy-based
optimization to learn explicit time-stepping operators governed by
free-energy functionals. \\
\hline

VB-DeepONet~\cite{garg2023vb} &
Computational mechanics &
Uses variational Bayesian inference to approximate the posterior
distribution of the network parameters and quantify predictive
uncertainty. \\
\hline

PI-DeepONet~\cite{tian2026physics} &
Structural mechanics &
Combines mixed-variable formulations with physics-informed operator
learning for real-time full-field prediction of stiffened-panel
bending responses under varying loading conditions. \\
\hline


\end{tabular}
\end{table*}

In addition to these advancements, recent studies have increasingly explored the application of DeepONet and physics-informed operator learning to structural dynamics and vibration analysis. Van Delden et al.~\cite{van2024learning} proposed an operator-learning-based framework for structural vibration prediction in harmonically excited plate structures. The proposed model effectively captured vibration responses across varying geometries and excitation conditions, outperforming DeepONet and the Fourier neural operator on a large-scale vibration benchmark. Ahmed et al.~\cite{ahmed2025physics} proposed a PI-DeepONet framework with stiffness-based loss functions for real-time structural response prediction. By incorporating equilibrium and energy conservation principles through structural stiffness matrices, the proposed approach achieved accurate displacement and rotation predictions with significantly reduced training time. The study demonstrated the potential of physics-informed operator learning for efficient structural analysis and surrogate modeling. Liu et al.~\cite{liu2024case} presented a preliminary study on vibration analysis using DeepONet for a single-degree-of-freedom (SDOF) vibration system. The authors demonstrated that DeepONet can effectively learn and predict structural vibration responses under various loading conditions. Their results highlighted the potential of operator learning frameworks for structural health monitoring and dynamic system analysis. Similarly, Wan et al.~\cite{wan2025deepvivonet} introduced DeepVIVONet, a DeepONet-based framework for dynamic reconstruction and forecasting of vortex-induced vibrations using sparse spatio-temporal measurements. The proposed model demonstrated accurate vibration prediction, transfer-learning capabilities, and effective sensor placement optimization for marine riser systems.

Overall, the existing literature demonstrates the growing capability of neural-operator frameworks for scientific computing and engineering applications. However, many existing approaches rely on paired input-output solution data or commonly employ uniformly distributed branch-network sensors. In particular, the use of CGL-distributed sensors within a physics-guided operator-learning framework for boundary-sensitive EBB vibration problems remains insufficiently explored. To address this gap, the proposed SpectONet combines physics-guided operator learning with CGL-based nonuniform sensor placement without requiring paired solution data. The next section introduces the benchmark EBB vibration problems considered in this work.

\section{Preliminaries}\label{sec_3}
This section provides the mathematical and theoretical background necessary for the proposed methodology. It first presents the problem statement, followed by an overview of the fundamental principles of PINNs and DeepONet.

 \subsection{Problem Statement}\label{Problem}\justifying

The transverse vibration of an EBB can be written as:~\cite{rao2019vibration}
\begin{equation}
\begin{aligned}
\rho A(x)\frac{\partial^2 u(x,t)}{\partial t^2}
&+ c(x)\frac{\partial u(x,t)}{\partial t} \\
&+ \frac{\partial^2}{\partial x^2}
\left[
EI(x)\frac{\partial^2 u(x,t)}{\partial x^2}
\right] \\
&= F(x,t),
\qquad x\in\Omega,\quad t\in[0,T],
\end{aligned}
\label{generalEBB}
\end{equation}
\noindent
where $u(x,t)$ denotes the transverse displacement of the beam,
$\rho(x)$ is the mass density of the beam material, $A(x)$ is the
cross-sectional area, $c(x)$ is the distributed viscous damping
coefficient, $EI(x)$ represents the spatially varying flexural rigidity,
and $F(x,t)$ denotes the external distributed load. The product
$\rho(x)A(x)$ represents the mass per unit length of the beam.
For spatially varying flexural rigidity, the operator
$\frac{\partial^2}{\partial x^2}
\left(EI(x)\frac{\partial^2u}{\partial x^2}\right)$
naturally accounts for the additional spatial derivative terms
arising from the variation of $EI(x)$. The symbols and physical
quantities used in the EBB model, together with their corresponding
units, are summarized in Table~\ref{notation}.

To obtain a unique solution, the following ICs are prescribed:
\begin{equation}
\left\{
\begin{aligned}
u(x,0) &= u_0(x),\\
\frac{\partial u}{\partial t}(x,0) &= v_0(x),
\end{aligned}
\right.
\qquad x\in\Omega.
\label{general_IC}
\end{equation}
where $u_0(x)$ and $v_0(x)$ represent the initial displacement and velocity distributions, respectively. The BCs of the considered EBB problems depend on the beam configuration. In general form, they can be expressed as:
\begin{equation}
\mathcal{B}_x[u(x,t)] = h(x,t),
\qquad x\in\partial\Omega,\; t\in[0,T],
\label{general_BC}
\end{equation}
where $\mathcal{B}_x[\cdot]$ denotes the differential boundary operator and $h(x,t)$ represents the specified boundary data.

\begin{figure*}
\centering
\includegraphics[width=0.75\textwidth]{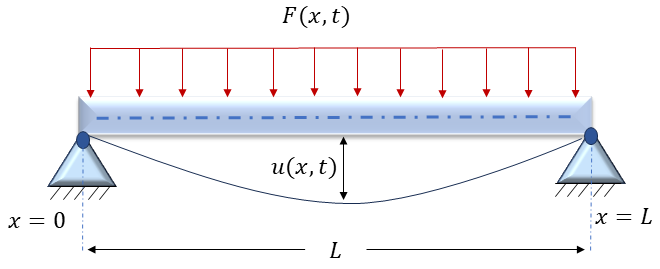}
\caption{Schematic representation of an Euler-Bernoulli
beam subjected to an external distributed transverse
load $F(x,t)$, illustrating the beam length $L$
and transverse displacement $u(x,t)$.}
\label{All_beam}
\end{figure*}

\begin{table}
\centering
\caption{Symbols and physical quantities employed in the EBB, together with their corresponding units.}
\label{notation}
\renewcommand{\arraystretch}{1.2}
\setlength{\tabcolsep}{3pt}

\begin{tabular}{p{0.14\columnwidth}
                p{0.58\columnwidth}
                p{0.18\columnwidth}}
\hline
\textbf{Symbol} & \textbf{Description} & \textbf{Unit} \\
\hline

$\rho$ &
Mass density of the beam material &
kg/m$^3$ \\

$A$ &
Cross-sectional area of the beam &
m$^2$ \\

$E$ &
Young's modulus of elasticity &
Pa \\

$I$ &
Area moment of inertia of the cross-section &
m$^4$ \\

$EI$ &
Flexural rigidity &
N$\cdot$m$^2$ \\

$c$ &
Distributed viscous damping coefficient  &
N$\cdot$s/m$^2$ \\

$u$ &
Transverse displacement &
m \\

$F$ &
External distributed transverse load &
N/m \\

\hline
\end{tabular}
\end{table}
 Eq.~(\ref{generalEBB}) provides a unified framework for
describing a wide range of structural vibration phenomena
encountered in EBB dynamics. By appropriately specifying
the external loading, BCs, and material parameters,
several classical beam vibration models can be obtained
as special cases. In the present study, three representative
EBB vibration problems are considered, namely, a forced
undamped beam with nonhomogeneous BCs,
an undamped simply supported beam with prescribed initial
displacement, and a harmonically forced undamped simply
supported beam initially at rest. A general schematic
representation of an EBB subjected to an external
distributed transverse load is illustrated in
Figure~\ref{All_beam}.


\begin{enumerate}

\item[(a)]
The forced undamped vibration of an EBB with
nonhomogeneous BCs is obtained
by setting $c(x)=0$ in Eq.~\eqref{generalEBB}.
The resulting fourth-order PDE is expressed as:
\begin{equation}
\begin{aligned}
\rho(x)A(x)\frac{\partial^2 u(x,t)}{\partial t^2}
&+
\frac{\partial^2}{\partial x^2}
\left[
EI(x)\frac{\partial^2 u(x,t)}{\partial x^2}
\right]\\
&=F(x,t),
\qquad x\in\Omega,\quad t\in[0,T].
\end{aligned}
\label{OL_undamped}
\end{equation}
The beam has a nonzero initial displacement
and zero initial velocity, prescribed as:
\begin{equation}
\left\{
\begin{aligned}
u(x,0)&=x^2(1-x)^2,\\
\frac{\partial u}{\partial t}(x,0)&=0,
\end{aligned}
\right.
\qquad x\in[0,1].
\label{P1_IC}
\end{equation}
The beam is subjected to homogeneous displacement
conditions and nonhomogeneous curvature conditions
at both ends, as specified below:
\begin{equation}
\left\{
\begin{aligned}
u(0,t)&=u(1,t)=0,\\
\frac{\partial^2u}{\partial x^2}(0,t)
&=2\cos t,\\
\frac{\partial^2u}{\partial x^2}(1,t)
&=2\cos t,
\end{aligned}
\right.
\qquad t\in[0,T].
\label{P1_BC}
\end{equation}

\item[(b)]
The transverse undamped vibration of a simply
supported EBB is also governed by
Eq.~\eqref{OL_undamped}. Unlike the preceding
configuration, homogeneous simply supported
BCs are imposed. The initial
displacement is prescribed as a sinusoidal
function, while the initial velocity is zero:
\begin{equation}
\left\{
\begin{aligned}
u(x,0)&=\sin(\pi x),\\
\frac{\partial u}{\partial t}(x,0)&=0,
\end{aligned}
\right.
\qquad x\in[0,1].
\label{P2_IC}
\end{equation}
The corresponding simply supported BCs are
expressed as:
\begin{equation}
\left\{
\begin{aligned}
u(0,t)&=u(1,t)=0,\\
\frac{\partial^2u}{\partial x^2}(0,t)
&=
\frac{\partial^2u}{\partial x^2}(1,t)=0,
\end{aligned}
\right.
\qquad t\in[0,T].
\label{P2_BC}
\end{equation}

\item[(c)]
The forced undamped vibration of an initially
resting simply supported EBB is governed by
Eq.~\eqref{OL_undamped}, with constant
material and geometric properties.
Since the beam is initially at rest,
the ICs are prescribed as: 
\begin{equation}
\left\{
\begin{aligned}
u(x,0)&=0,\\
\frac{\partial u}{\partial t}(x,0)&=0,
\end{aligned}
\right.
\qquad x\in[0,1].
\label{P3_IC}
\end{equation}
The corresponding simply supported BCs are
identical to those prescribed in
Eq.~\eqref{P2_BC}.
\end{enumerate}

\subsection{Physics-informed Neural Networks}

PINNs introduced by Raissi et al.~\cite{raissi2019physics} are a class of deep learning models for solving forward and inverse problems governed by PDEs. Unlike conventional data-driven neural networks, PINNs incorporate the governing PDE together with the associated ICs and BCs into the loss function, thereby embedding physical laws directly into the training process~\cite{ketkar2021automatic}.

For the EBB vibration problem governed by
Eq.~\eqref{generalEBB}, the transverse displacement $u(x,t)$ is
approximated by a neural network $u_\theta(x,t)$, where $\theta$
denotes the trainable network parameters. Using AD, the physics residual corresponding to
Eq.~\eqref{generalEBB} is defined as:
\begin{equation}
\begin{aligned}
R_\theta(x,t)
={}&
\rho A(x)
\frac{\partial^2 u_\theta(x,t)}{\partial t^2}
+
c(x)
\frac{\partial u_\theta(x,t)}{\partial t}
\\
&+
\frac{\partial^2}{\partial x^2}
\left[
EI(x)
\frac{\partial^2 u_\theta(x,t)}{\partial x^2}
\right]
-
F(x,t).
\end{aligned}
\label{eq:pinn_residual}
\end{equation}
The trainable parameters are obtained by minimizing
the composite loss function
\begin{equation}
\mathcal{L}(\theta)
=
\lambda_r\mathcal{L}_{\mathrm{PDE}}
+
\lambda_i\mathcal{L}_{\mathrm{IC}}
+
\lambda_b\mathcal{L}_{\mathrm{BC}},
\label{total_loss}
\end{equation}
where $\lambda_r$, $\lambda_i$, and $\lambda_b$
denote the weighting coefficients associated with
the PDE residual, IC, and BC losses, respectively.
In the numerical implementation, the BCs and
ICs losses are further decomposed into
their individual constraints, allowing separate
weighting of each loss component.
\noindent
The PDE residual loss is defined as:
\begin{equation}
\mathcal{L}_{\mathrm{PDE}}
=
\frac{1}{N_r}
\sum_{i=1}^{N_r}
\left|
R_\theta(x_i^r,t_i^r)
\right|^2,
\label{eq:pinn_pde_loss}
\end{equation}

According to the ICs prescribed in
Eq.~\eqref{general_IC}, the IC loss consists of the initial
displacement and initial velocity constraints and is expressed as:
\begin{equation}
\begin{aligned}
\mathcal{L}_{\mathrm{IC}}
={}&
\frac{1}{N_0}
\sum_{i=1}^{N_0}
\Bigg[
\left|
u_\theta(x_i^0,0)-u_0(x_i^0)
\right|^2
\\
&+
\left|
\frac{\partial u_\theta}{\partial t}(x_i^0,0)
-
v_0(x_i^0)
\right|^2
\Bigg].
\end{aligned}
\label{eq:pinn_ic_loss}
\end{equation}

According to the BCs prescribed in
Eq.~\eqref{general_BC}, the corresponding BC loss is defined as:
\begin{equation}
\mathcal{L}_{\mathrm{BC}}
=
\frac{1}{N_b}
\sum_{i=1}^{N_b}
\left|
\mathcal{B}_x[u_\theta](x_i^b,t_i^b)
-
h(x_i^b,t_i^b)
\right|^2,
\label{eq:pinn_bc_loss}
\end{equation}
where $N_r$, $N_0$, and $N_b$ denote the numbers of interior,
initial, and boundary collocation points, respectively, with
$\{(x_i^r,t_i^r)\}_{i=1}^{N_r}$,
$\{x_i^0\}_{i=1}^{N_0}$, and
$\{(x_i^b,t_i^b)\}_{i=1}^{N_b}$ representing their corresponding
collocation-point sets.
\noindent
Finally, the optimal network parameters are obtained by solving
\begin{equation}
\theta^{*}
=
\arg\min_{\theta}
\mathcal{L}(\theta).
\label{eq:pinn_opt}
\end{equation}

Although PINNs have demonstrated remarkable success in solving a broad range of PDEs, they remain solution-specific, requiring retraining whenever the governing equation, physical parameters, forcing function, or IC and BC change. Furthermore, the computational cost can become significant for complex and many-query problems. These limitations have motivated the development of operator learning approaches, such as DeepONet, which learn mappings between function spaces rather than individual PDE solutions.

\subsection{Deep Operator Networks}
\justifying

DeepONet, proposed by Lu et al.~\cite{lu2021learning}, is a neural
operator architecture designed to learn nonlinear mappings between
infinite-dimensional function spaces. Unlike conventional neural
networks, which approximate mappings between finite-dimensional
vectors, DeepONet directly approximates nonlinear operators. The
input function may represent quantities such as an IC, BC, source
term, or material property. Owing to the universal approximation
(Theorem~\ref{Theorem_operator}), DeepONet provides an effective
framework for learning solution operators associated with PDEs.

For the EBB vibration problem defined in Eq.~\eqref{generalEBB},
the objective of DeepONet is to learn the solution operator
\begin{equation}
\mathcal{G}:g(x)\mapsto u(x,t),
\label{eq:solution_operator}
\end{equation}
where $g(x)$ denotes the spatial component of the prescribed
external forcing function, and $u(x,t)$ represents the
corresponding transverse displacement field under the specified
ICs and BCs. The temporal dependence of the external forcing
is prescribed for each considered problem. The trained DeepONet
approximates the solution operator as
$\widehat{\mathcal{G}}_{\theta}(g)(x,t)$, where $\theta=(W,b)$
denotes the trainable weights and biases. Its branch network
encodes the sampled forcing function, while the trunk network
encodes the spatio-temporal coordinates.

Since the branch network operates on finite-dimensional inputs,
the continuous forcing function is first discretized by sampling
it at a finite set of prescribed sensor locations, denoted by
$\{x_1,x_2,\ldots,x_m\}$. The resulting discretized input is
represented as:
\begin{equation}
\mathbf{g}
=
\left[
g(x_1),
g(x_2),
\ldots,
g(x_m)
\right]^T,
\label{eq:branch_input}
\end{equation}
which serves as the input feature vector for the branch network.
The trunk network receives the spatio-temporal coordinates
$(x,t)\in\Omega\times\mathcal{T}$. The branch and trunk networks
generate the latent feature vectors
$\mathbf{b}=[\hat{b}_1,\hat{b}_2,\ldots,\hat{b}_p]^T$ and
$\mathbf{t}=[\hat{t}_1,\hat{t}_2,\ldots,\hat{t}_p]^T$,
respectively, where both vectors have the same latent dimension $p$.
The DeepONet approximation is then obtained through their inner
product,
\begin{equation}
\widehat{\mathcal{G}}_{\theta}(g)(x,t)
=
\sum_{k=1}^{p}
\hat{b}_k(\mathbf{g})\,\hat{t}_k(x,t),
\label{eq:deeponet_operator}
\end{equation}
where $\hat{b}_k(\mathbf{g})$ denotes the latent feature extracted
by the branch network from the discretized forcing function,
whereas $\hat{t}_k(x,t)$ represents the basis function generated
by the trunk network corresponding to the spatio-temporal
coordinates.

\begin{theorem}[Universal approximation theorem for operators~\cite{lu2021learning}]
\label{Theorem_operator}
Let $\sigma$ be a continuous non-polynomial activation function.
Let $X$ be a Banach space, and let $K_1\subset X$ and
$K_2\subset\mathbb{R}^{d}$ be compact sets. Assume that $V$ is a
compact subset of $C(K_1)$ and that
$\mathcal{G}:V\rightarrow C(K_2)$ is a continuous nonlinear
operator. Then, for every $\varepsilon>0$, there exist positive
integers $m$, $n$, and $p$, sensor locations $x_j\in K_1$, scalar
coefficients $c_i^k,\ \xi_{ij}^k,\ \theta_i^k,\ \zeta_k\in
\mathbb{R}$, and vectors $w_k\in\mathbb{R}^{d}$ such that
\begin{equation}
\begin{aligned}
\Bigg|
\mathcal{G}(g)(y)
-
\sum_{k=1}^{p}
&\left(
\sum_{i=1}^{n}
c_i^k
\sigma\left(
\sum_{j=1}^{m}
\xi_{ij}^{k}g(x_j)+\theta_i^k
\right)
\right)
\\
&\times
\sigma\left(w_k\cdot y+\zeta_k\right)
\Bigg|
<\varepsilon ,
\end{aligned}
\label{eq:universal_operator}
\end{equation}
for all $g\in V$ and $y\in K_2$.
\end{theorem}

The above theorem establishes the mathematical foundation of
DeepONet by demonstrating its ability to approximate a broad class
of continuous nonlinear operators with arbitrary accuracy.

For supervised training, let the dataset consist of $N_f$ input
forcing functions, each associated with $N_q$ spatio-temporal
query points and their corresponding reference displacement
values. The training dataset is defined as:
\begin{equation}
\begin{aligned}
\mathcal{D}=\Big\{
&\big(g^{(i)},\mathbf{y}_j^{(i)},
u^{(i)}(\mathbf{y}_j^{(i)})\big)
:\\
&i=1,\ldots,N_f,\quad
j=1,\ldots,N_q
\Big\},
\end{aligned}
\label{eq:deeponet_dataset}
\end{equation}
where $\mathbf{y}_j^{(i)}=(x_j^{(i)},t_j^{(i)})$
denotes the $j$th spatio-temporal query point associated
with the $i$th input forcing function, and
$u^{(i)}(\mathbf{y}_j^{(i)})
=\mathcal{G}(g^{(i)})(\mathbf{y}_j^{(i)})$
represents the corresponding reference displacement.
The supervised DeepONet loss function is defined as:
\begin{equation}
\begin{aligned}
\mathcal{L}_{\mathrm{DeepONet}}
={}&\frac{1}{N_fN_q}
\sum_{i=1}^{N_f}
\sum_{j=1}^{N_q}
\Big|
\widehat{\mathcal{G}}_{\theta}
(g^{(i)})(\mathbf{y}_j^{(i)})
\\
&\qquad -
u^{(i)}(\mathbf{y}_j^{(i)})
\Big|^{2}.
\end{aligned}
\label{eq:deeponet_loss}
\end{equation}

Although DeepONet is capable of accurately learning nonlinear
operators, its original formulation does not explicitly incorporate
the PDE and associated physical constraints into the learning
process. Consequently, its performance may deteriorate when
training data are limited or when strong physical consistency
is required. These limitations motivate the development of
physics-guided operator learning frameworks that embed governing
equations into the optimization process, leading to improved
accuracy, enhanced data efficiency, and physically consistent
predictions.


\section{Proposed Model}\label{sec_4}\justifying

\begin{figure*}
\centering
\includegraphics[width=\textwidth]{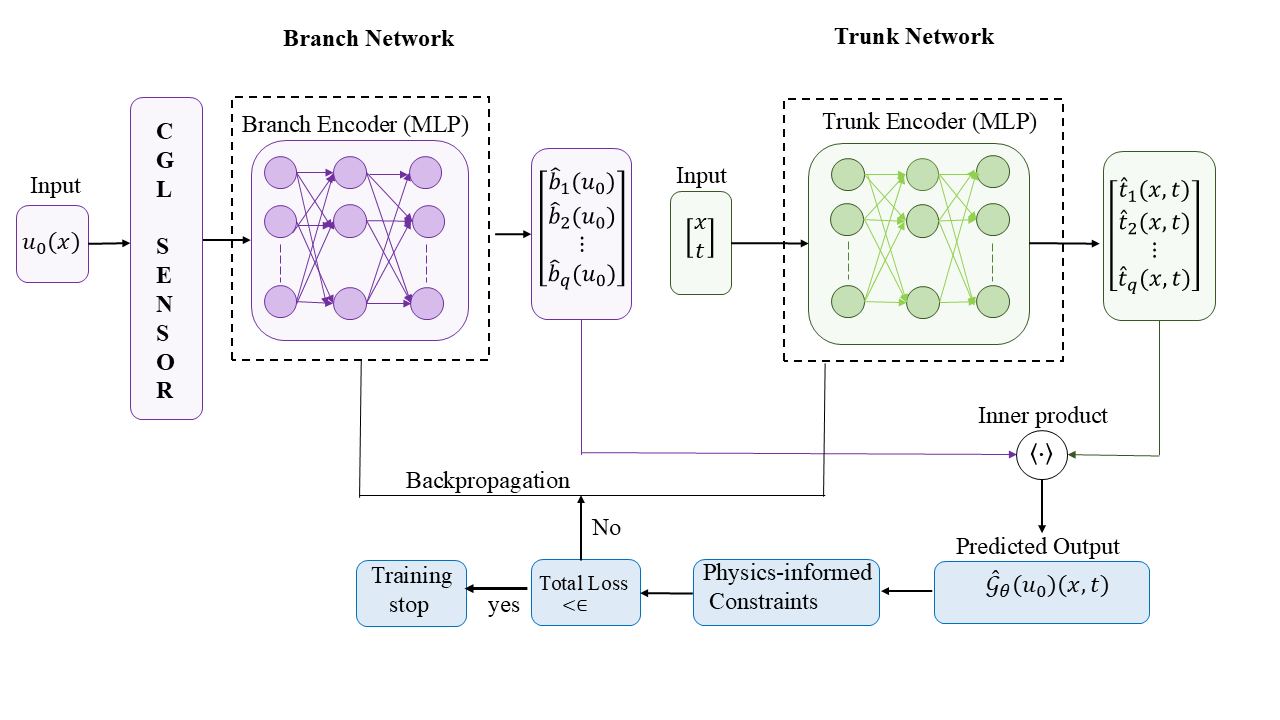}
\caption{Architecture and training workflow of the SpectONet framework for solving EBB vibration problems.}
\label{SpectONet}
\end{figure*}

The SpectONet architecture shown in Figure~\ref{SpectONet}
consists of two interconnected multilayer perceptrons (MLPs):
a branch network and a trunk network. For the generalized EBB
vibration model given in Eq.~\eqref{generalEBB}, the objective
is to learn the solution operator
\begin{equation}
\mathcal{G}:g\mapsto u,
\qquad
\mathcal{G}(g)(x,t)=u(x,t),
\label{operator_mapping}
\end{equation}
where $g(x)$ denotes the spatial component of the prescribed
external forcing function, and $u(x,t)$ is the corresponding
spatio-temporal displacement field for the specified ICs and BCs.

In SpectONet, the branch network encodes $g(x)$ through its
CGL sensor representation, while the trunk network encodes
the spatio-temporal coordinates $(x,t)$ at which the beam
displacement is evaluated. The learned representations from
the branch and trunk networks are combined using an
inner-product operation to obtain the final neural network
prediction.

\subsection{Branch Network with CGL Sensors Representation}

The branch network is responsible for learning the dependence of the solution on the input forcing function \(g(x)\). Since \(g(x)\) is a continuous function, it cannot be directly supplied to a neural network. Therefore, it is first converted into a finite-dimensional vector by evaluating it at a set of CGL sensor locations.
\noindent
The CGL nodes are derived from the extrema of the Chebyshev polynomials of the first kind, defined as:~\cite{el2016chebyshev}
\begin{equation}
T_N(\xi)
=
\cos\left(N\arccos \xi\right),
\qquad
\xi\in[-1,1].
\label{chebyshev}
\end{equation}
The extrema of \(T_N(\xi)\), or equivalently the roots of
\begin{equation}
(1-\xi^2)T_N'(\xi)=0,
\label{cgl_root}
\end{equation}
define the CGL nodes on the reference interval \([-1,1]\) as:
\begin{equation}
\xi_j
=
-\cos\left(\frac{j\pi}{N}\right),
\qquad
j=0,1,\ldots,N.
\label{cgl_reference}
\end{equation}

\noindent
Since the EBB problem is considered over the physical domain \(\Omega=[0,1]\), the reference CGL nodes are mapped to the physical domain as:
\begin{equation}
x_j
=
\frac{\xi_j+1}{2}
=
\frac{1}{2}
\left[
1-
\cos\left(\frac{j\pi}{N}\right)
\right],
\qquad
j=0,1,\ldots,N.
\label{physical_nodes}
\end{equation}
\noindent
Using these CGL sensor locations, the continuous input forcing function \(g(x)\) is represented by the sensor vector
\begin{equation}
\mathbf{g}
=
\left[
g(x_0),
g(x_1),
\ldots,
g(x_N)
\right]^T
\in \mathbb{R}^{N+1}.
\label{branch_input_vector}
\end{equation}
This CGL-based sensor vector is supplied to the branch network as the finite-dimensional representation of the input forcing function.

Unlike uniformly distributed sensors, the CGL sensors naturally cluster near both boundaries of the computational domain and remain relatively sparse in the interior. To illustrate this boundary clustering, let
$\theta_j=\frac{j\pi}{N}.$ For \(j\ll N\), the Taylor expansion
\[
\cos\theta
=
1-\frac{\theta^2}{2}
+\mathcal{O}\!\left(\theta^4\right)
\]
gives
\begin{align}
x_j
&=
\frac{1-\cos\theta_j}{2}
\nonumber\\
&=
\frac{\pi^2j^2}{4N^2}
+
\mathcal{O}\left(\frac{j^4}{N^4}\right)
\nonumber\\
&=
\mathcal{O}\left(\frac{j^2}{N^2}\right),
\qquad j\ll N.
\label{boundary_cluster}
\end{align}
Consequently, the spacing between two consecutive sensors satisfies
\begin{align}
\Delta x_j
:=
x_{j+1}-x_j
&=
\frac{\pi^2(2j+1)}{4N^2}
+
\mathcal{O}\!\left(\frac{(j+1)^3}{N^4}\right)
\nonumber\\
&=
\mathcal{O}\!\left(\frac{j+1}{N^2}\right),
\qquad j\ll N.
\label{spacing}
\end{align}

Thus, the CGL sensor spacing becomes increasingly small near the boundaries. Beyond this boundary clustering, CGL nodes also provide a more stable distribution for polynomial approximation than uniformly spaced nodes. In particular, the corresponding Lebesgue constant grows logarithmically with the number of nodes,
\[
\Lambda_N=\mathcal{O}(\log N),
\]
whereas the Lebesgue constant associated with uniformly spaced nodes may grow much more rapidly. Consequently, CGL nodes limit the amplification of interpolation errors and mitigate the endpoint oscillations associated with the Runge phenomenon. However, these approximation-theoretic properties do not by themselves guarantee improved neural-network prediction accuracy.

The resulting CGL-based sensor vector $\mathbf{g}$
is supplied to the branch network to learn a latent
representation of the input forcing function. Following the
DeepONet framework~\cite{lu2021learning}, the branch
network is implemented as an MLP consisting of fully connected layers. The
network processes the sampled function values through
successive affine transformations and nonlinear
activation functions. The feature representation
at the $l^{th}$ hidden layer is defined as:
\begin{equation}
\mathbf{h}_b^{(l)}
=
\phi_b
\left(
\mathbf{W}_b^{(l)}
\mathbf{h}_b^{(l-1)}
+
\mathbf{c}_b^{(l)}
\right),
\qquad
l=1,2,\ldots,L_b,
\label{branch_hidden}
\end{equation}
where the initial branch input is given by
\begin{equation}
\mathbf{h}_b^{(0)}
=
\mathbf{g}
\in\mathbb{R}^{N+1}.
\label{branch_initial}
\end{equation}
Here, $\mathbf{h}_b^{(l)}\in\mathbb{R}^{d_l}$
denotes the feature vector at the $l^{th}$ hidden
layer, $\mathbf{W}_b^{(l)}
\in\mathbb{R}^{d_l\times d_{l-1}}$ and
$\mathbf{c}_b^{(l)}\in\mathbb{R}^{d_l}$
represent the trainable weight matrix and
bias vector, respectively, with $d_0=N+1$.
Furthermore, $\phi_b(\cdot)$ denotes the
nonlinear activation function, and $L_b$
is the number of hidden layers.

The output of the final hidden layer is mapped
to a $p$-dimensional latent space through a
fully connected output layer. Let
$\mathcal{B}_{\theta_b}$ denote the branch
network parameterized by $\theta_b$. The
resulting latent representation is expressed as
\begin{equation}
\mathcal{B}_{\theta_b}(\mathbf{g})
=
\left[
\hat{b}_1(\mathbf{g}),
\hat{b}_2(\mathbf{g}),
\ldots,
\hat{b}_p(\mathbf{g})
\right]^T
\in\mathbb{R}^{p},
\label{branch_output_vector}
\end{equation}
where $q$ denotes the latent dimension and
$\hat{b}_k(\mathbf{g})$, $k=1,2,\ldots,p$,
represents the $k^{th}$ learned feature of the
input forcing function. The resulting latent features
are subsequently combined with the corresponding
trunk network outputs to construct the
predicted solution operator.

\subsection{Trunk Network}

The trunk network learns the dependence of the solution on the spatio-temporal coordinates. It takes the coordinate pair \((x,t)\) as input and maps it into the same \(p\)-dimensional output space as the branch network. Let \(\mathcal{T}_{\theta_t}\) denote the trunk network with trainable parameters \(\theta_t\). For an \(L_t\)-layer fully connected trunk network, the hidden layers are defined as:
\begin{equation}
\mathbf{h}_t^{(l)}
=
\phi_t
\left(
\mathbf{W}_t^{(l)}\mathbf{h}_t^{(l-1)}
+
\mathbf{c}_t^{(l)}
\right),
\qquad
l=1,2,\ldots,L_t,
\label{trunk_hidden}
\end{equation}
where the initial trunk input is given by
\begin{equation}
\mathbf{h}_t^{(0)}
=
[x,t]^T
\in\mathbb{R}^{2}.
\label{trunk_initial}
\end{equation}
Here, \(\mathbf{W}_t^{(l)}\) and \(\mathbf{c}_t^{(l)}\) denote the trainable weight matrix and bias vector of the $l^{th}$ trunk layer, respectively, and \(\phi_t(\cdot)\) denotes the nonlinear activation function used in the trunk network.
\noindent
The output of the final hidden layer is projected to the trunk output vector
\begin{equation}
\mathcal{T}_{\theta_t}(x,t)
=
\left[
\hat{t}_1(x,t),\,
\hat{t}_2(x,t),\,
\ldots,\,
\hat{t}_p(x,t)
\right]^T
\in\mathbb{R}^{p}.
\label{trunk_output_vector}
\end{equation}
Here, \(p\) denotes the output dimension of the trunk network, and each component \(\hat{t}_k(x,t)\), \(k=1,2,\ldots,p\), is a learned coordinate-dependent feature associated with the spatio-temporal input \((x,t)\).

After obtaining the branch and trunk output vectors, SpectONet combines them through an inner-product operation. The branch network provides the learned features associated with the input forcing function, while the trunk network provides the learned coordinate-dependent features. The learned operator \(\widehat{\mathcal{G}}_{\theta}\) approximates the exact solution operator \(\mathcal{G}\), and its output is expressed as:
\begin{equation}
\widehat{\mathcal{G}}_{\theta}(g)(x,t)
=
\mathcal{B}_{\theta_b}(\mathbf{g})
\cdot
\mathcal{T}_{\theta_t}(x,t)
=
\sum_{k=1}^{p}
\hat{b}_k(\mathbf{g})\hat{t}_k(x,t).
\label{learned_operator_output}
\end{equation}

\subsection{Loss Function}

To enforce the governing physics, AD~\cite{baydin2018automatic}
is employed to compute the required temporal and spatial
derivatives of the predicted operator output
$\widehat{\mathcal{G}}_{\theta}(g^{(i)})(x,t)$~\cite{wang2021learning}.
During training, the branch network receives a collection
of sampled spatial forcing functions
$\{g^{(i)}\}_{i=1}^{N_f}$, where $N_f$ denotes the number
of training input functions used to learn the solution operator.
For the $i^{th}$ sampled forcing function, the network
prediction is given by
\[
\widehat{\mathcal{G}}_{\theta}\!\left(g^{(i)}\right)(x,t)
\approx
\mathcal{G}\!\left(g^{(i)}\right)(x,t)
=
u^{(i)}(x,t),
\]
where $u^{(i)}(x,t)$ denotes the corresponding displacement
solution for the sampled forcing function $g^{(i)}(x)$.
The external
forcing associated with each input function is denoted by
$F^{(i)}(x,t)$, whose temporal dependence is prescribed
for the considered benchmark problem.

\begin{figure*}
\centering
\includegraphics[width=0.62\textwidth]{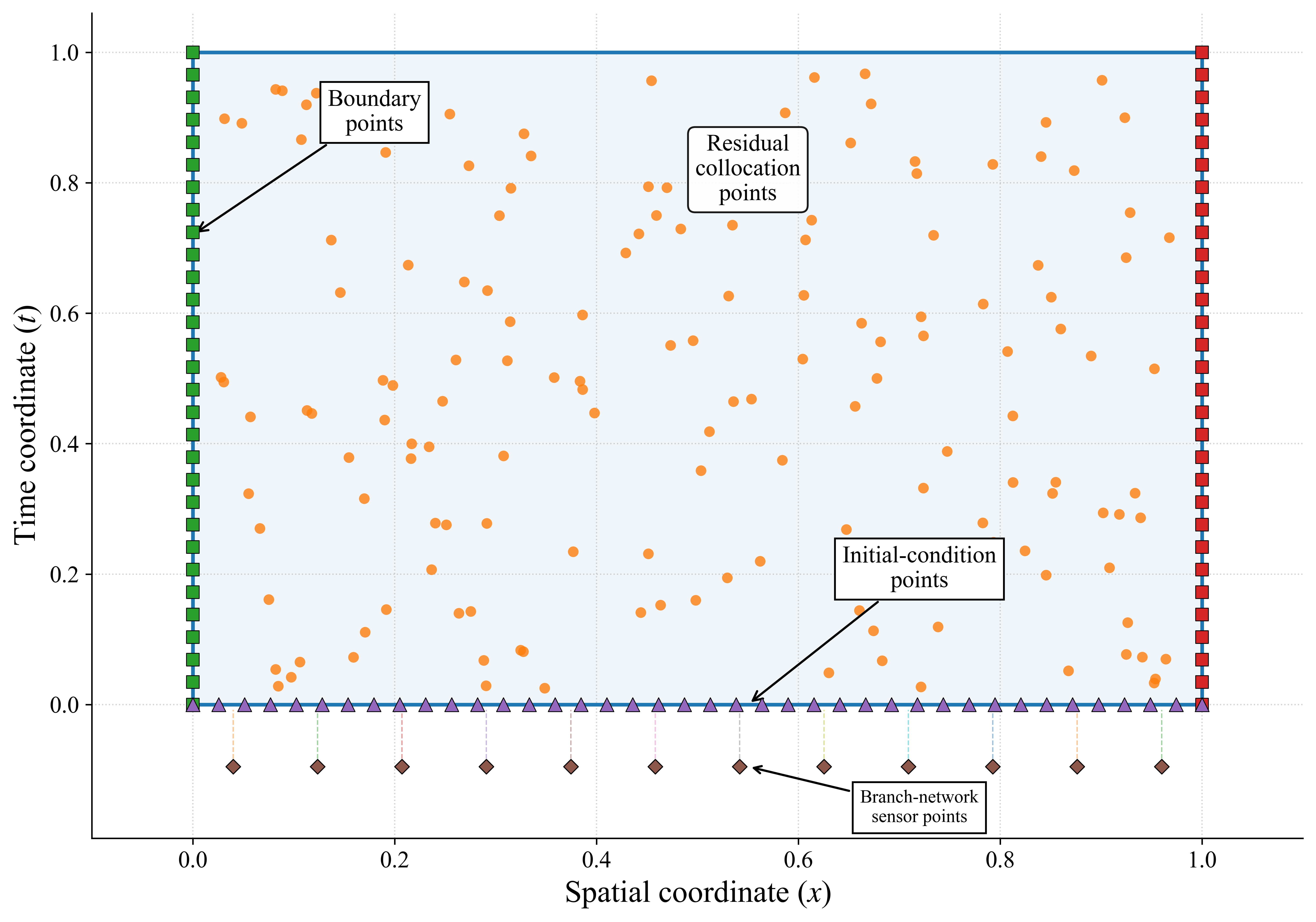}
\caption{Training domain in the space-time plane for SpectONet. The interior collocation points are sampled inside the domain, while the boundary points are taken along \(x=0\) and \(x=1\), and the initial points are sampled at \(t=0\).}
\label{fig:training_domain}
\end{figure*}

For each sampled input function, the governing equation is
enforced at interior collocation points, while the prescribed
ICs and BCs are imposed at the respective training points.
Accordingly, the physics-informed residual corresponding
to Eq.~\eqref{generalEBB} is defined as:
\begin{equation}
\begin{aligned}
R_{\theta}\!\left(g^{(i)}\right)(x,t)
={}&\;
\rho A(x)
\frac{\partial^2
\widehat{\mathcal{G}}_{\theta}
\!\left(g^{(i)}\right)(x,t)}
{\partial t^2}
\\
&+
c(x)
\frac{\partial
\widehat{\mathcal{G}}_{\theta}
\!\left(g^{(i)}\right)(x,t)}
{\partial t}
\\
&+
\frac{\partial^2}{\partial x^2}
\left[
EI(x)
\frac{\partial^2
\widehat{\mathcal{G}}_{\theta}
\!\left(g^{(i)}\right)(x,t)}
{\partial x^2}
\right]
\\
&-
F^{(i)}(x,t).
\end{aligned}
\label{residual_SpectONet}
\end{equation}
\noindent
Here, $R_{\theta}$ measures the extent to which the
predicted solution satisfies the governing EBB equation.
The corresponding loss functions are given by

\begin{align}
\mathcal{L}_{PDE}
={}&\;
\frac{1}{N_fN_r}
\sum_{i=1}^{N_f}
\sum_{j=1}^{N_r}
\left|
R_{\theta}\!\left(g^{(i)}\right)
(x_j^r,t_j^r)
\right|^2,
\label{loss_pde_spectonet}
\\[1ex]
\mathcal{L}_{BC}
={}&\;
\frac{1}{N_fN_b}
\sum_{i=1}^{N_f}
\sum_{j=1}^{N_b}
\left|
\mathcal{B}_x
\!\left[
\widehat{\mathcal{G}}_{\theta}
\!\left(g^{(i)}\right)
(x_j^b,t_j^b)
\right]
\right.
\nonumber\\
&\left.
\qquad
-h(x_j^b,t_j^b)
\right|^2,
\label{loss_bc_spectonet}
\\[1ex]
\mathcal{L}_{IC}
={}&\;
\frac{1}{N_fN_0}
\sum_{i=1}^{N_f}
\sum_{j=1}^{N_0}
\left|
\widehat{\mathcal{G}}_{\theta}
\!\left(g^{(i)}\right)(x_j^0,0)
\right.
\nonumber\\
&\left.
\qquad
-u_0(x_j^0)
\right|^2
\nonumber\\
&+
\frac{1}{N_fN_0}
\sum_{i=1}^{N_f}
\sum_{j=1}^{N_0}
\left|
\frac{\partial}{\partial t}
\widehat{\mathcal{G}}_{\theta}
\!\left(g^{(i)}\right)(x_j^0,0)
\right.
\nonumber\\
&\left.
\qquad
-v_0(x_j^0)
\right|^2,
\label{loss_ic_spectonet}
\end{align}
where $\mathcal{L}_{PDE}$, $\mathcal{L}_{BC}$, and
$\mathcal{L}_{IC}$ denote the PDE residual, BC, and IC
losses, respectively. The initial displacement $u_0(x)$
and initial velocity $v_0(x)$ are prescribed for each
benchmark problem and remain fixed across the sampled
forcing functions. Moreover, $N_f$ denotes the number
of sampled input forcing functions, while $N_r$, $N_b$,
and $N_0$ represent the numbers of interior collocation
points, boundary points, and initial points associated
with each sampled input function. The distribution of
these training points over the space-time domain is
illustrated in Figure~\ref{fig:training_domain}. The overall training objective is obtained by combining
the PDE residual, IC, and BC losses according to
Eq.~(\ref{total_loss}). The optimal network parameters
are obtained by solving Eq.~(\ref{eq:pinn_opt}).

The principal novelty of the proposed framework lies in the seamless integration of the CGL sensors representation with a physics-guided operator learning architecture. Consequently, the proposed framework unifies mathematically motivated sensor placement, operator learning, and physics-guided training within a single architecture, resulting in improved accuracy for EBB vibration analysis.

\begin{algorithm}[t]
\caption{Training procedure of the proposed SpectONet framework}
\label{alg:FLDeepONet}
\SetAlgoLined

Initialize branch parameters $\theta_b$, trunk parameters $\theta_t$,
and set $\theta=\{\theta_b,\theta_t\}$\;

Construct the CGL sensor locations
$x_j=\dfrac{1}{2}\left(1-\cos\left(\dfrac{j\pi}{N}\right)\right),
\; j=0,1,\ldots,N$\;

Generate the training input forcing functions
$\{g^{(i)}(x)\}_{i=1}^{N_f}$\;

\For{$i=1,2,\ldots,N_f$}{
    Construct the corresponding CGL sensor vector
    $\mathbf{g}^{(i)}
    =[g^{(i)}(x_0),g^{(i)}(x_1),\ldots,
    g^{(i)}(x_N)]^T$\;
}

Generate interior collocation points $(x_j^r,t_j^r)$,
boundary points $(x_j^b,t_j^b)$,
and initial points $(x_j^0,0)$\;

Set convergence threshold $\varepsilon$\;

\For{each Adam training epoch}{
    \For{each mini-batch of indexed input sensor vectors
    $\{\mathbf{g}^{(i)}\}$}{

        Compute branch features using
        Eq.~\eqref{branch_output_vector}\;

        Compute trunk features using
        Eq.~\eqref{trunk_output_vector}\;

        Predict the solution using
        Eq.~\eqref{learned_operator_output}\;

        Compute the PDE residual
        $R_\theta(g^{(i)})(x,t)$
        from Eq.~\eqref{residual_SpectONet} using AD\;

        Compute the PDE residual loss $\mathcal{L}_{PDE}$
        using Eq.~\eqref{loss_pde_spectonet}\;

        Compute the BC loss $\mathcal{L}_{BC}$
        using Eq.~\eqref{loss_bc_spectonet}\;

        Compute the IC loss $\mathcal{L}_{IC}$
        using Eq.~\eqref{loss_ic_spectonet}\;

        Compute the total loss according to
        Eq.~\eqref{total_loss}\;

        Update $\theta$ using the Adam optimizer\;
    }
}

Refine the network parameters $\theta$ using the
L-BFGS optimizer by minimizing the total loss
defined in Eq.~\eqref{total_loss}\;

\Return{Trained SpectONet model}\;

\end{algorithm}

\section{Numerical Experiments}\label{sec_5}\justifying

This section presents the numerical experiments performed to evaluate the proposed SpectONet framework for solving EBB vibration problems. The performance of the proposed model is compared with four deep learning baseline models, namely PI-DeepONet~\cite{cong2026respecting}, Vanilla DeepONet~\cite{lu2021learning}, PINNs~\cite{raissi2019physics}, and CNN-UNet~\cite{ronneberger2015u}, in terms of prediction accuracy, convergence behavior, and error distribution. Detailed description of the architectures, training configurations, and implementation settings of these baseline models are provided in Appendix ~\ref{Apndx}. Three benchmark problems are considered, namely
Problem~1 (P1): forced undamped EBB with nonhomogeneous
BCs, Problem~2 (P2): forced undamped
simply supported EBB,
and Problem~3 (P3): forced undamped initially resting
simply supported EBB.

\subsection{Error Metrics}

To quantitatively assess the predictive accuracy of the proposed framework,
the relative $\mathcal{L}_2$ error, root mean square error (RMSE), mean
absolute error (MAE), and maximum absolute error (Max Error) are employed.
Let $u^{(i)}(x,t)$ and
$\widehat{\mathcal{G}}_{\theta}(g^{(i)})(x,t)$ denote the exact and
predicted solutions, respectively, corresponding to the $i^{th}$ sampled
spatial forcing function $g^{(i)}(x)$. Suppose the solutions are evaluated
at $N$ points
$\{(x_j,t_j)\}_{j=1}^{N}$.
The error metrics are defined as follows:

\begin{equation}
\text{Relative } \mathcal{L}_2 \text{ error}
=
\sqrt{
\frac{
\displaystyle\sum_{j=1}^{N}
\left(
u^{(i)}(x_j,t_j)
-
\widehat{\mathcal{G}}_{\theta}(g^{(i)})(x_j,t_j)
\right)^2
}{
\displaystyle\sum_{j=1}^{N}
\left(
u^{(i)}(x_j,t_j)
\right)^2
}
}.
\label{eq:relative_l2_error}
\end{equation}

\begin{equation}
\mathrm{RMSE}
=
\sqrt{
\frac{1}{N}
\sum_{j=1}^{N}
\left(
u^{(i)}(x_j,t_j)
-
\widehat{\mathcal{G}}_{\theta}(g^{(i)})(x_j,t_j)
\right)^2
}.
\label{eq:rmse}
\end{equation}

\begin{equation}
\mathrm{MAE}
=
\frac{1}{N}
\sum_{j=1}^{N}
\left|
u^{(i)}(x_j,t_j)
-
\widehat{\mathcal{G}}_{\theta}(g^{(i)})(x_j,t_j)
\right|.
\label{eq:mae}
\end{equation}

\begin{equation}
\mathrm{Max\ Error}
=
\max_{1\le j\le N}
\left|
u^{(i)}(x_j,t_j)
-
\widehat{\mathcal{G}}_{\theta}(g^{(i)})(x_j,t_j)
\right|.
\label{eq:max_error}
\end{equation}

\begin{equation}
\mathrm{Improvement (\%)}
=
\frac{
\mathrm{RMSE}_{\mathrm{baseline}}
-
\mathrm{RMSE}_{\mathrm{SpectONet}}
}{
\mathrm{RMSE}_{\mathrm{baseline}}
}
\times100\%.
\end{equation}

\subsection{Experimental Setup}
All numerical experiments were performed under the same computational environment to ensure a fair comparison among the considered models. The hardware and software configuration used throughout this study is summarized in Table~\ref{system_setup}.

The network architectures and optimization hyperparameters employed for the proposed SpectONet are summarized in Table~\ref{SpectONet_parameters}. Unless otherwise specified, the same optimization strategy and training settings are adopted across all benchmark problems, while only the branch and trunk network architectures are adjusted according to the complexity of each problem.

\begin{table}
\centering
\caption{Hardware and software configuration used for the numerical experiments.}
\label{system_setup}
\begin{tabularx}{\columnwidth}{@{}l X@{}}
\toprule
Component & Specification \\
\midrule
Processor & AMD Ryzen 7 5700U with Radeon Graphics \\
CPU Configuration & 8 cores / 16 threads \\
System Memory & 8 GB RAM (7.35 GB usable) \\
Computing Platform & CPU-based computation \\
Operating System & Windows 11 (64-bit) \\
Programming Language & Python 3.11.1 \\
Deep Learning Framework & PyTorch 2.7.1 \\
\bottomrule
\end{tabularx}
\end{table}

\begin{table*}
\centering
\caption{Network architecture and hyperparameters employed for the proposed model.}
\label{SpectONet_parameters}
\renewcommand{\arraystretch}{1.15}
\begin{tabular}{lccc}
\toprule
\textbf{Parameter} & \textbf{P1} & \textbf{P2} & \textbf{P3}\\
\midrule

Activation function & Tanh & Tanh & Tanh\\
Learning rate & $1e-03$ & $2e-03$ & $1e-02$\\
Adam iterations & 500 & 500 & 3000\\
L-BFGS iterations & 5000 & 3000 & 5000\\
Evaluation grid & $101\times101$ & $101\times101$ & $101\times101$\\
Branch network architecture & $[12,10,10]$ & $[12,24,24]$ & $[12,24,24]$\\
Trunk network architecture & $[2,10,10]$ & $[2,24,24]$ & $[2,24,24]$\\
Loss weights
& $[1,10,10,10,10,10,10]$
& $[1,10,10,10,10,10,10]$
& $[1,10,10,10,10,10,10]$ \\

\bottomrule
\end{tabular}
\end{table*}

\subsection{Forced Undamped Euler-Bernoulli Beam with Nonhomogeneous Boundary Conditions}
\label{P1}

For the first benchmark problem (P1), we consider the forced undamped vibration of an EBB with nonhomogeneous BCs. The proposed SpectONet is employed to learn the solution operator that maps a prescribed external forcing function to the corresponding transverse displacement field.

For this problem, the material and geometric parameters in Eq.~\eqref{OL_undamped} are specified as~\cite{aouragh2024compact}
\[
\rho A(x)=1,\qquad EI(x)=1,
\]
corresponding to an undamped beam with constant coefficients. The external forcing function is assumed to have the separable form
\begin{equation}
F(x,t)=g(x)\cos t,
\end{equation}
where $g(x)$ denotes the spatial component of the distributed load. Substituting these quantities into Eq.~\eqref{OL_undamped}, the governing equation reduces to
\begin{equation}
\frac{\partial^2u}{\partial t^2}
+
\frac{\partial^4u}{\partial x^4}
=
g(x)\cos t,
\qquad x\in[0,1],\;t\in[0,1].
\label{P1_PDE}
\end{equation}

\noindent The corresponding ICs and nonhomogeneous BCs are prescribed in Eqs.~\eqref{P1_IC} and \eqref{P1_BC}, respectively.

\noindent
For the forcing function
\begin{equation}
g(x)=24-x^2(1-x)^2,
\end{equation}
the corresponding analytical solution is given by~\cite{aouragh2024compact}
\begin{equation}
u(x,t)=x^2(1-x)^2\cos t.
\label{P1_exact}
\end{equation}
This analytical solution is used as a reference for evaluating the predictive accuracy of SpectONet for the specified forcing function.

Thus, the operator learning problem is expressed as
\begin{equation}
\widehat{\mathcal{G}}_{\theta}: g(x)\mapsto u(x,t),
\end{equation}
where $\widehat{\mathcal{G}}_{\theta}$ denotes the neural operator approximated by the SpectONet. The branch network receives the values of the spatial forcing function $g(x)$ at selected sensor locations, while the trunk network takes the spatio-temporal coordinates $(x,t)$ as input. The output of the network approximates the displacement response $u(x,t)$ over the computational domain. The network architecture and training parameters used for this problem are summarized in Table~\ref{SpectONet_parameters}.

In this study, the branch input is constructed using 12 CGL sensor locations over the beam domain $[0,1]$. Taking $N=11$ in Eq.~(\ref{physical_nodes}) yields a total of $N+1=12$ sensor points. Consequently, the branch network receives the discrete forcing representation

\begin{equation}
\mathbf{g}
=
[g(x_0),g(x_1),\ldots,g(x_{11})]^T
\in
\mathbb{R}^{12}.
\end{equation}

These CGL sensor points are clustered near the two beam boundaries, thereby improving the representation of boundary-dominated vibration features without increasing the number of sensors.
\\
The PDE residual associated with P1 is defined as:

\begin{equation}
\begin{aligned}
R_{\theta}\!\left(g^{(i)}\right)(x,t)
={}&
\frac{\partial^2\widehat{\mathcal{G}}_{\theta}
(g^{(i)})(x,t)}{\partial t^2}
\\
&+
\frac{\partial^4\widehat{\mathcal{G}}_{\theta}
(g^{(i)})(x,t)}{\partial x^4}
-
g^{(i)}(x)\cos t.
\end{aligned}
\end{equation}
The residual loss is given by
\begin{equation}
\begin{aligned}
\mathcal{L}_{\mathrm{PDE}}
={}&
\frac{1}{N_fN_r}
\sum_{i=1}^{N_f}
\sum_{j=1}^{N_r}
\Bigg|
\frac{\partial^2
\widehat{\mathcal{G}}_{\theta}(g^{(i)})(x_j^r,t_j^r)}
{\partial t^2}
\\
&+
\frac{\partial^4
\widehat{\mathcal{G}}_{\theta}(g^{(i)})(x_j^r,t_j^r)}
{\partial x^4}
-
g^{(i)}(x_j^r)\cos(t_j^r)
\Bigg|^2,
\end{aligned}
\end{equation}
where $N_r$ denotes the number of collocation points sampled inside the computational domain. The initial displacement loss is defined as:

\begin{equation}
\begin{aligned}
\mathcal{L}_{\mathrm{IC}}
={}&
\frac{1}{N_fN_0}
\sum_{i=1}^{N_f}
\sum_{j=1}^{N_0}
\Bigg[
\left|
\widehat{\mathcal{G}}_{\theta}(g^{(i)})
(x_j^0,0)
-
(x_j^0)^2(1-x_j^0)^2
\right|^2
\\
&+
\left|
\frac{\partial
\widehat{\mathcal{G}}_{\theta}(g^{(i)})
(x_j^0,0)}
{\partial t}
\right|^2
\Bigg].
\end{aligned}
\end{equation}
The BC loss is introduced to enforce the displacement and bending moment conditions and is defined as:

\begin{equation}
\begin{aligned}
\mathcal{L}_{\mathrm{BC}}
={}&
\frac{1}{N_fN_b}
\sum_{i=1}^{N_f}
\sum_{j=1}^{N_b}
\Bigg[
\left|
\widehat{\mathcal{G}}_{\theta}(g^{(i)})
(0,t_j^b)
\right|^2
\\
&+
\left|
\widehat{\mathcal{G}}_{\theta}(g^{(i)})
(1,t_j^b)
\right|^2
\\
&+
\left|
\frac{\partial^2
\widehat{\mathcal{G}}_{\theta}(g^{(i)})
(0,t_j^b)}
{\partial x^2}
-
2\cos t_j^b
\right|^2
\\
&+
\left|
\frac{\partial^2
\widehat{\mathcal{G}}_{\theta}(g^{(i)})
(1,t_j^b)}
{\partial x^2}
-
2\cos t_j^b
\right|^2
\Bigg].
\end{aligned}
\end{equation}
The overall training objective is obtained by combining the residual, IC, and BC losses as defined in Eq.~\eqref{total_loss}.

 To enforce the governing equation together with the associated ICs and BCs, 128 initial points, 128 boundary points, and 1024 interior collocation points are randomly sampled from the spatio-temporal domain $(x,t)\in[0,1]\times[0,1]$. The proposed SpectONet is trained using a two-stage optimization strategy, where the Adam optimizer is first applied for 500 iterations to obtain an initial approximation, followed by the L-BFGS optimizer for a maximum of 5000 iterations to further refine the network parameters.

\begin{figure*}
\centering
\setlength{\tabcolsep}{6pt}
\renewcommand{\arraystretch}{1.5}

\fbox{
\begin{tabular}{c|c}
\textbf{Predicted Solutions} &
\textbf{Absolute Error}  \\
\hline
\\[-1mm]
\includegraphics[width=0.47\textwidth]{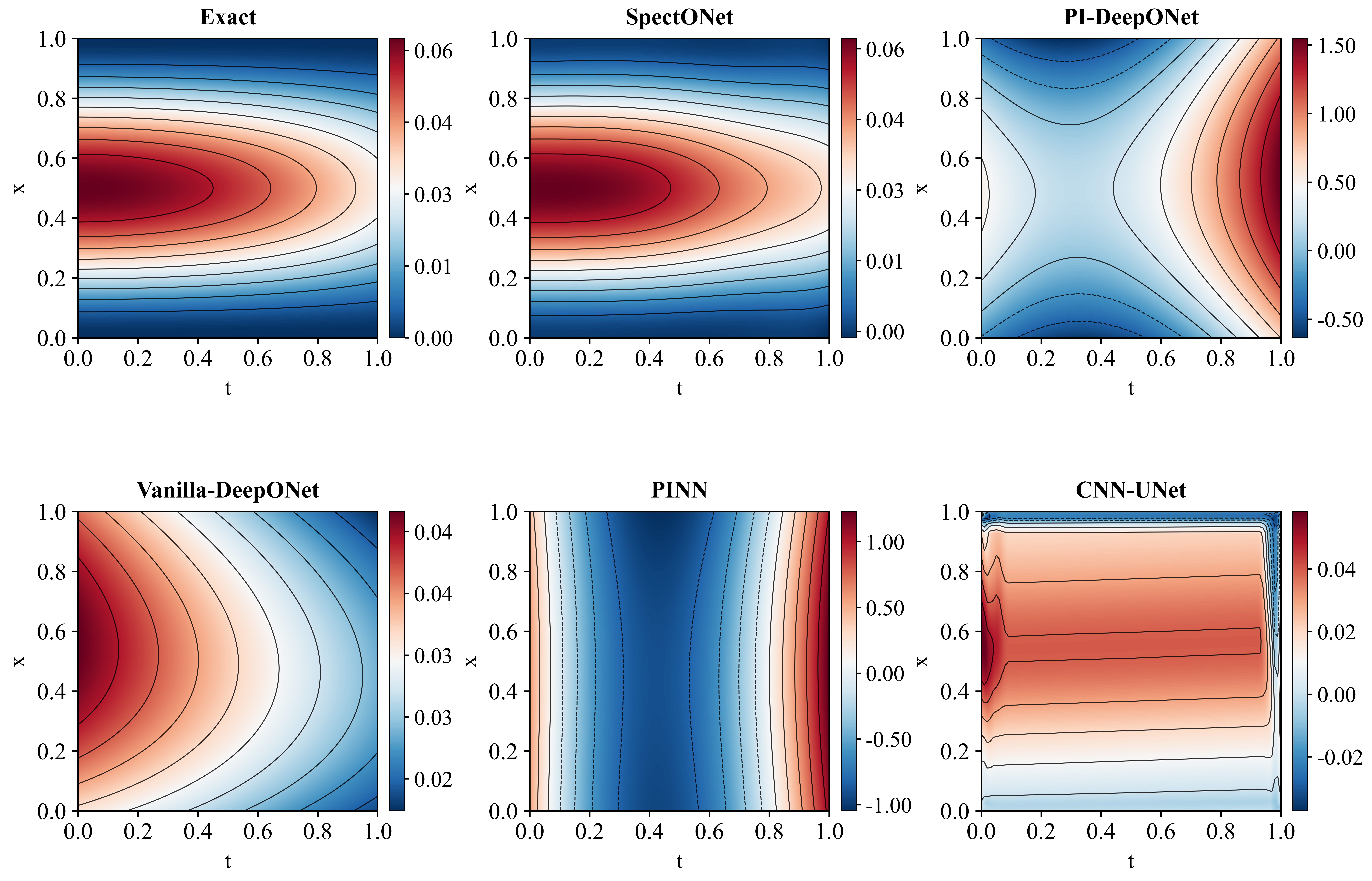} &
\includegraphics[width=0.47\textwidth]{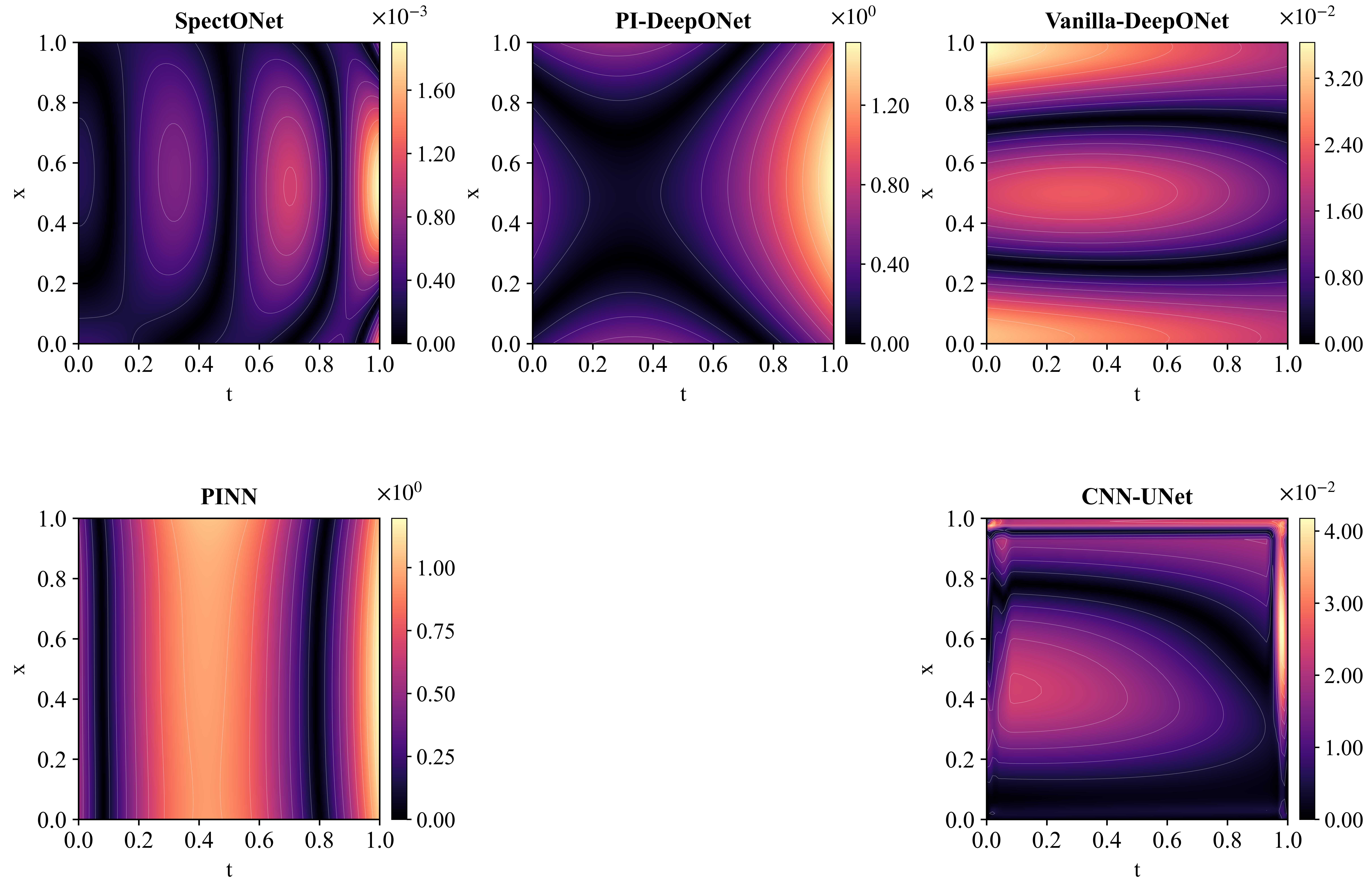} \\

\end{tabular}
}

\caption{
Qualitative comparison of the proposed SpectONet with the baseline models for P1. The left column presents the 2D predicted solution fields, whereas the right column shows the corresponding point-wise absolute error distributions over the spatio-temporal domain $(x,t)\in[0,1]\times[0,1].$
}
\label{fig:P1_box}
\end{figure*}

Figure~\ref{fig:P1_box} presents the two-dimensional (2D) predicted displacement fields and their corresponding point-wise absolute error distributions for SpectONet and the baseline models over the spatio-temporal domain $(x,t)\in[0,1]\times[0,1]$. The results demonstrate that SpectONet closely reproduces the displacement patterns of the exact solution, with minimal visible discrepancies throughout the computational domain. In contrast, PI-DeepONet and PINN exhibit substantial deviations from the reference solution, whereas Vanilla-DeepONet and CNN-UNet provide comparatively improved approximations but still show noticeable discrepancies. The corresponding absolute error distributions further confirm that SpectONet produces substantially smaller errors than the baseline models.

\begin{figure*}
    \centering
    \includegraphics[width=0.8\textwidth]{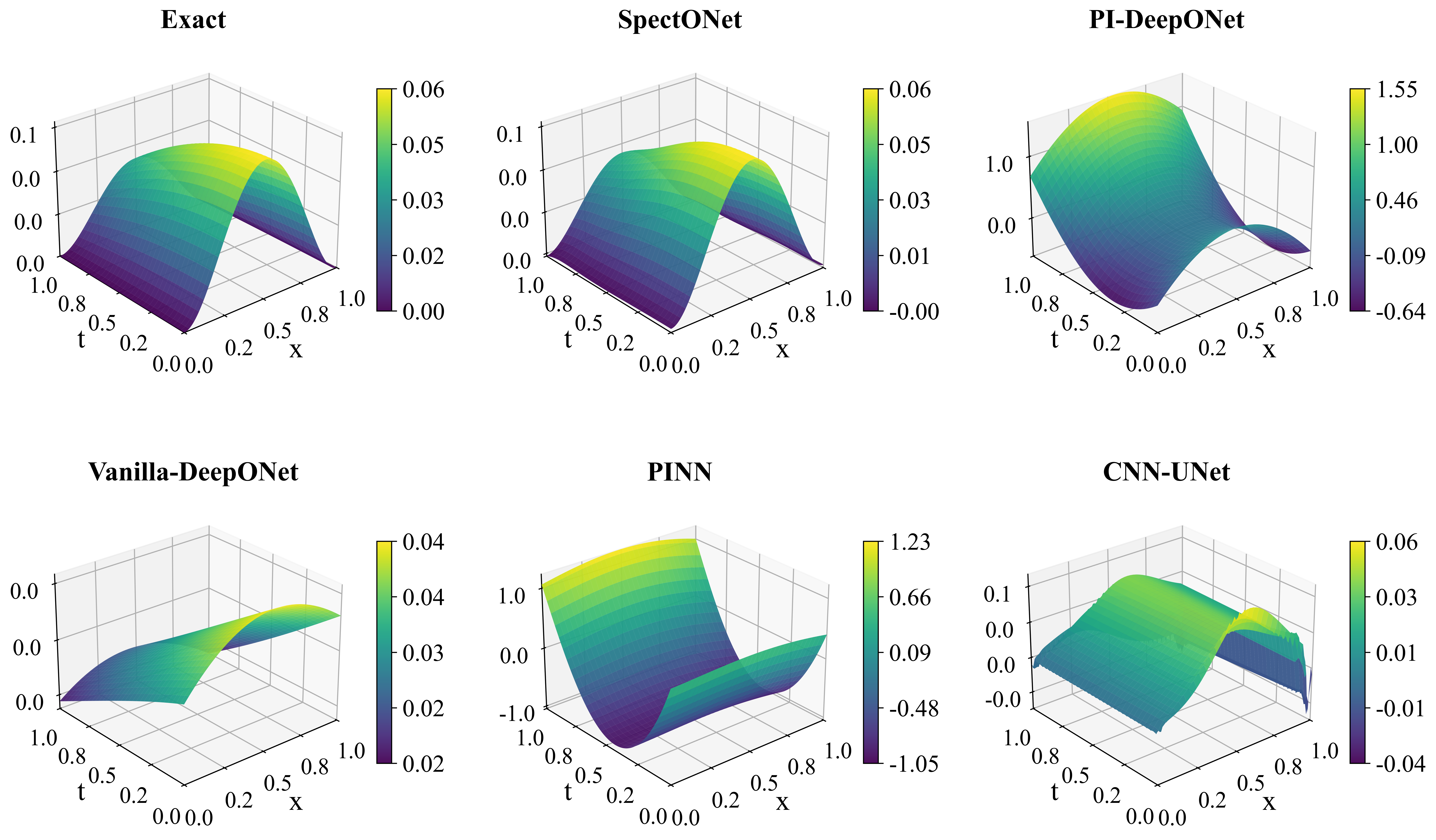}
   \caption{3D comparison of the exact solution and the predicted solutions obtained by the proposed SpectONet for P1 over the spatio-temporal domain $(x,t)\in[0,1]\times[0,1].$}
   \label{3d_Solution_comparison_P1}
\end{figure*}

A further comparison of the three-dimensional (3D) solution
surfaces obtained using the proposed SpectONet and the
baseline models is presented in
Figure~\ref{3d_Solution_comparison_P1}. The predicted surface of SpectONet closely agrees with the exact solution, accurately capturing the spatial and temporal variations in beam displacement. In comparison, the baseline models exhibit varying degrees of deviation from the analytical solution, consistent with the observations from the 2D displacement fields.

Table~\ref{tab:problem1_errors} reports the
quantitative performance of SpectONet and four
baseline models across four independent random
seeds, denoted by S1 (1234), S2 (42), S3 (2024),
and S4 (321). The relative $\mathcal{L}_2$ error,
RMSE, MAE, and P95 error are reported as mean
values with their corresponding standard deviations,
whereas the maximum error is reported as the mean
across the four seeds. The P95 error denotes the
95th percentile of the point-wise absolute errors,
indicating that 95\% of the evaluation points have
absolute errors at or below this value.
SpectONet achieves the lowest mean error across all
reported evaluation metrics, with a mean relative
$\mathcal{L}_2$ error of $2.831e-{2}$. This corresponds
to reductions of approximately 94.21\%, 99.84\%, 99.85\%,
and 96.56\% compared with Vanilla-DeepONet,
PI-DeepONet, PINN, and CNN-UNet, respectively.


\begin{table*}
\centering
\caption{Performance comparison of the proposed model with the baseline models for P1 (mean $\pm$ standard deviation over four seeds, S1-S4).}
\label{tab:problem1_errors}
\renewcommand{\arraystretch}{1.2}
\setlength{\tabcolsep}{3pt}

\resizebox{\textwidth}{!}{%
\begin{tabular}{lcccccc}
\hline
\textbf{Model} &
\textbf{Rel. $\mathcal{L}_2$ Error} &
\textbf{RMSE} &
\textbf{MAE} &
\textbf{P95 Error} &
\textbf{Max. Error} \\
\hline

\textbf{SpectONet} &
$\mathbf{2.831e{-2}\pm1.536e{-2}}$ &
$\mathbf{9.567e{-4}\pm5.191e{-4}}$ &
$\mathbf{7.564e{-4}\pm4.029e{-4}}$ &
$\mathbf{1.955e{-3}\pm1.110e{-3}}$ &
$\mathbf{3.437e{-3}}$ \\

PI-DeepONet &
$1.744e{+1}\pm2.167e{+0}$ &
$5.892e{-1}\pm7.322e{-2}$ &
$4.806e{-1}\pm7.054e{-2}$ &
$1.168e{+0}\pm2.308e{-1}$ &
$1.444e{+0}$  \\

Vanilla-DeepONet &
$4.886e{-1}\pm6.902e{-2}$ &
$1.651e{-2}\pm2.332e{-3}$ &
$1.448e{-2}\pm2.066e{-3}$ &
$2.718e{-2}\pm3.997e{-3}$ &
$3.362e{-2}$  \\

PINN &
$1.914e{+1}\pm9.811e{+0}$ &
$6.466e{-1}\pm3.315e{-1}$ &
$5.619e{-1}\pm2.889e{-1}$ &
$1.077e{+0}\pm5.615e{-1}$ &
$1.267e{+0}$  \\

CNN-UNet &
$8.235e{-1}\pm7.988e{-1}$ &
$2.783e{-2}\pm2.699e{-2}$ &
$2.503e{-2}\pm2.595e{-2}$ &
$4.197e{-2}\pm3.452e{-2}$ &
$5.469e{-2}$ \\

\hline
\end{tabular}%
}
\end{table*}

\begin{figure*}
    \centering
    \includegraphics[width=0.95\textwidth]{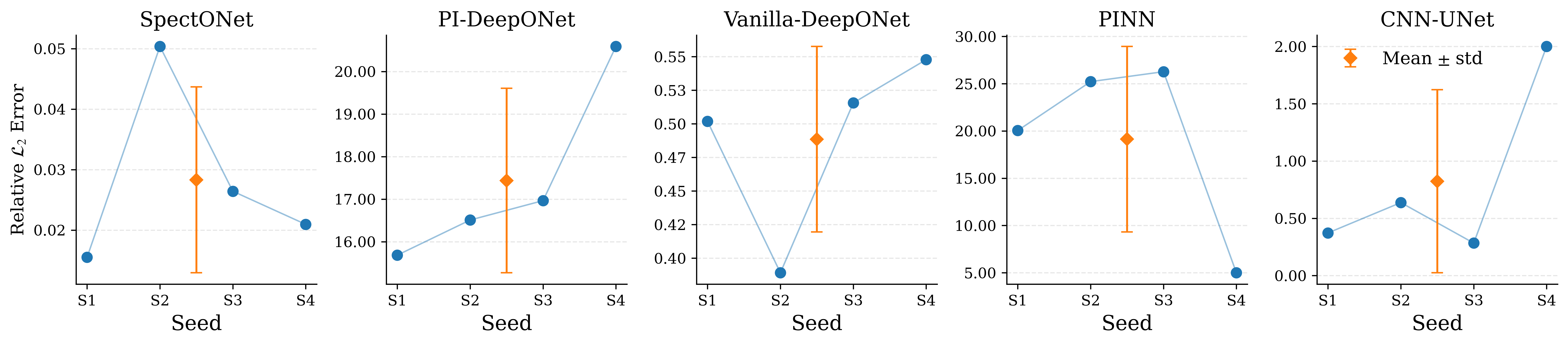}
    \caption{Comparison of the relative $\mathcal{L}_2$ errors of SpectONet and four baseline models across four independent random seeds S1-S4 for P1.}
    \label{fig:p1_seed_robustness}
\end{figure*}

To further investigate the influence of random
seeds on predictive performance,
Figure~\ref{fig:p1_seed_robustness} compares the
relative $\mathcal{L}_2$ errors of SpectONet and
the four baseline models across S1-S4.
SpectONet consistently achieves lower relative
errors than the competing models across all four
seeds. These observations further demonstrate
the predictive accuracy and robustness of
SpectONet across different random seeds.

\subsection{Forced Undamped Simply Supported Euler-Bernoulli Beam}
\label{P2}

The second benchmark problem (P2) considers an undamped simply supported EBB, as illustrated in Figure~\ref{All_beam}. The material and geometric parameters in Eq.~\eqref{OL_undamped} are specified as:~\cite{aouragh2024compact}
\[
\rho A(x)=1,\qquad EI(x)=1,
\]
corresponding to a beam with constant coefficients. The external forcing function is assumed to have the separable form
\begin{equation}
F(x,t)=g(x)\cos t,
\end{equation}
where $g(x)$ denotes the spatial distribution of the applied load. Substituting these quantities into Eq.~\eqref{OL_undamped} yields
\begin{equation}
\frac{\partial^2u}{\partial t^2}
+
\frac{\partial^4u}{\partial x^4}
=
g(x)\cos t,
\qquad x\in[0,1],\;t\in[0,1].
\label{P2_PDE}
\end{equation}
Unlike P1, this benchmark considers a sinusoidal initial displacement and homogeneous simply supported BCs. The corresponding ICs and BCs are prescribed in Eqs.~\eqref{P2_IC} and \eqref{P2_BC}, respectively. 
The analytical solution corresponding to the forcing function $g(x)=(\pi^{4}-1)\sin(\pi x)$ is given by~\cite{aouragh2024compact}
\begin{equation}
u(x,t)=\sin(\pi x)\cos t.
\end{equation}
Following the operator-learning formulation introduced in P1, the PDE residual loss for P2 is defined as:
\begin{equation}
\begin{aligned}
\mathcal{L}_{\mathrm{PDE}}
={}&
\frac{1}{N_fN_r}
\sum_{i=1}^{N_f}
\sum_{j=1}^{N_r}
\Bigg|
\frac{\partial^2
\widehat{\mathcal{G}}_{\theta}\!\left(g^{(i)}\right)
(x_j^r,t_j^r)}
{\partial t^2}
\\
&\quad+
\frac{\partial^4
\widehat{\mathcal{G}}_{\theta}\!\left(g^{(i)}\right)
(x_j^r,t_j^r)}
{\partial x^4}
-
g^{(i)}(x_j^r)\cos(t_j^r)
\Bigg|^2 .
\end{aligned}
\end{equation}
The IC loss is defined as:
\begin{equation}
\begin{aligned}
\mathcal{L}_{\mathrm{IC}}
={}&
\frac{1}{N_fN_0}
\sum_{i=1}^{N_f}
\sum_{j=1}^{N_0}
\Bigg[
\left|
\widehat{\mathcal{G}}_{\theta}(g^{(i)})
(x_j^0,0)
-
\sin(\pi x_j^0)
\right|^2
\\
&+
\left|
\frac{\partial
\widehat{\mathcal{G}}_{\theta}(g^{(i)})
(x_j^0,0)}
{\partial t}
\right|^2
\Bigg].
\end{aligned}
\label{eq:IC_loss}
\end{equation}
The boundary loss is introduced to enforce the simply supported conditions and is given by

\begin{equation}
\begin{aligned}
\mathcal{L}_{\mathrm{BC}}
={}&
\frac{1}{N_fN_b}
\sum_{i=1}^{N_f}
\sum_{j=1}^{N_b}
\Bigg[
\left|\widehat{\mathcal{G}}_{\theta}(g^{(i)})(0,t_j^b)\right|^2
+
\left|\widehat{\mathcal{G}}_{\theta}(g^{(i)})(1,t_j^b)\right|^2
\\
&+
\left|
\frac{\partial^2\widehat{\mathcal{G}}_{\theta}(g^{(i)})(0,t_j^b)}
{\partial x^2}
\right|^2
+
\left|
\frac{\partial^2\widehat{\mathcal{G}}_{\theta}(g^{(i)})(1,t_j^b)}
{\partial x^2}
\right|^2
\Bigg].
\end{aligned}
\label{eq:BC_loss}
\end{equation}

\noindent
The overall training objective is obtained by combining the residual, IC, and BC losses as defined in Eq.~\eqref{total_loss}.

The network architecture and problem-specific training
hyperparameters adopted for P2 are summarized in
Table~\ref{SpectONet_parameters}. To enforce the governing
equation together with the prescribed ICs and BCs,
48 initial points, 48 boundary points, and 256 interior
collocation points are randomly sampled from the
spatio-temporal domain $(x,t)\in[0,1]\times[0,1]$.

\begin{figure*}
\centering
\setlength{\tabcolsep}{6pt}
\renewcommand{\arraystretch}{1.5}

\fbox{
\begin{tabular}{c|c}
\textbf{Predicted Solutions} &
\textbf{Absolute Error} \\ 
\hline
\\[-1mm]
\includegraphics[width=0.47\textwidth]{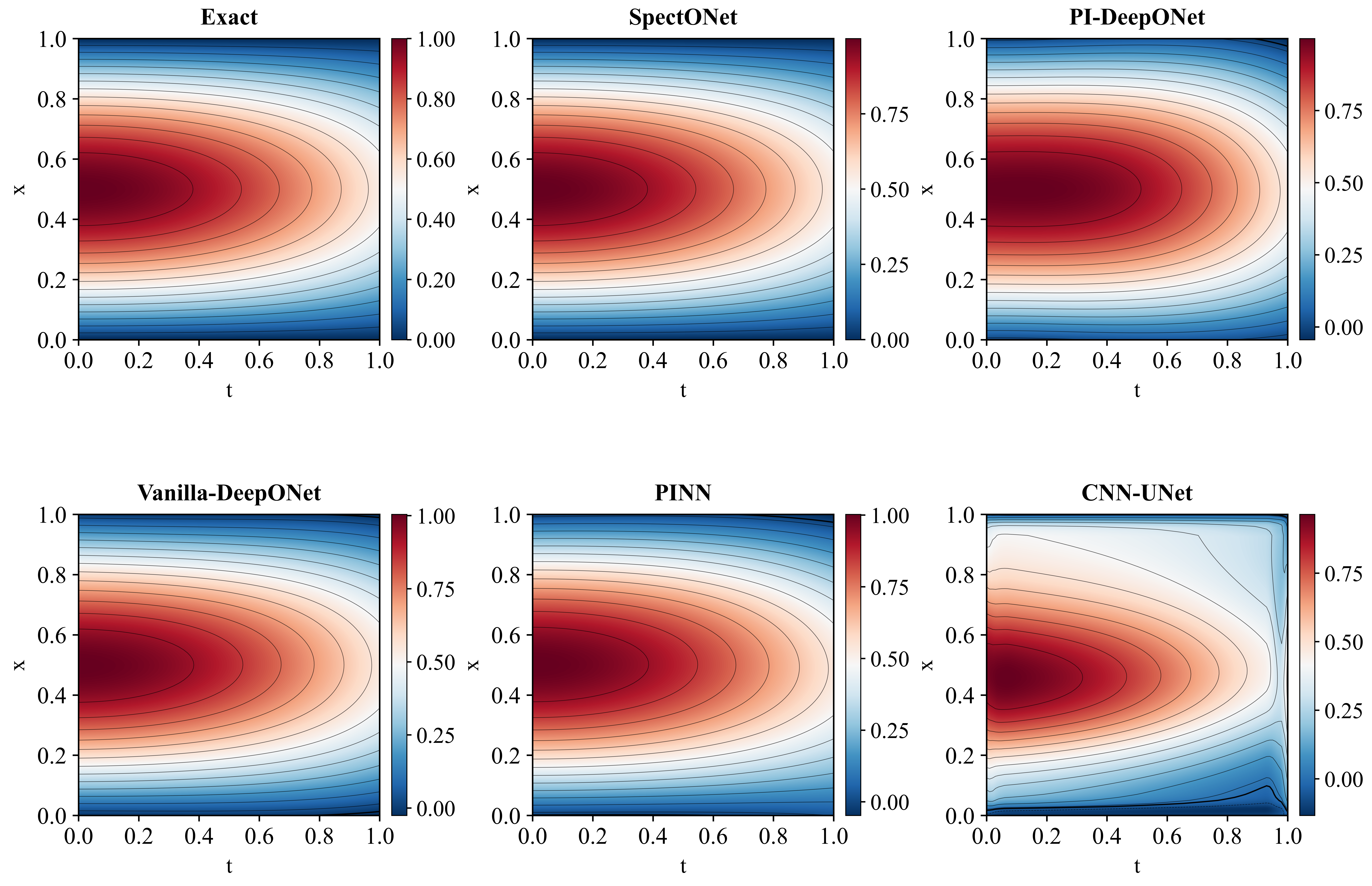} &
\includegraphics[width=0.47\textwidth]{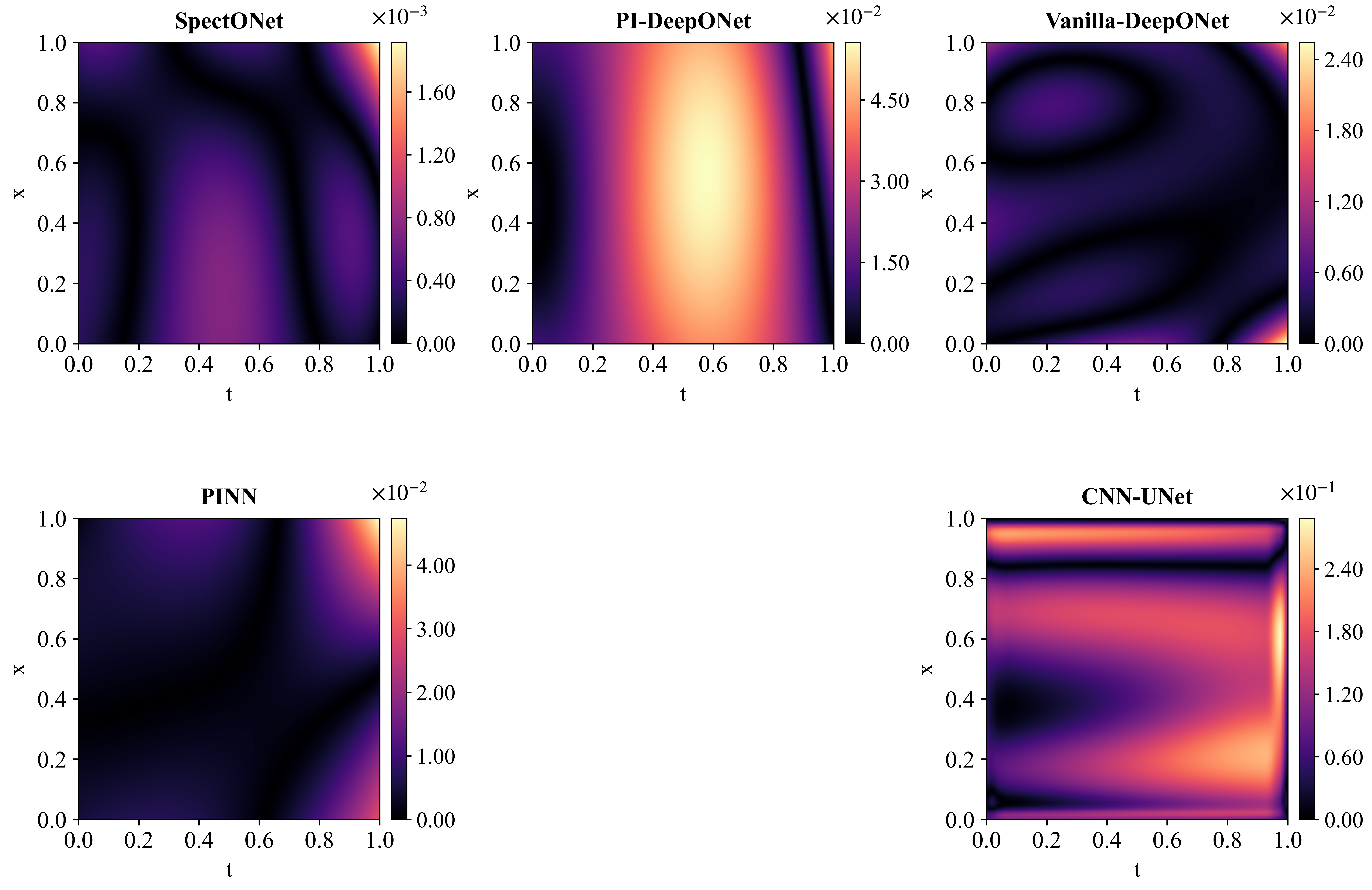} \\

\end{tabular}
}

\caption{
Qualitative comparison of the proposed SpectONet with the baseline models for P2. The left column presents the 2D predicted solution fields, whereas the right column shows the corresponding point-wise absolute error distributions over the spatio-temporal domain $(x,t)\in[0,1]\times[0,1].$
}
\label{fig:P2_box}
\end{figure*}

The spatial and temporal displacement profiles obtained for P2 are illustrated in Figure~\ref{fig:P2_box}, together with the associated point-wise absolute errors. SpectONet captures the sinusoidal vibration response with close agreement to the analytical reference, maintaining low prediction errors across the computational domain. The PI-DeepONet and PINN predictions exhibit pronounced discrepancies, while Vanilla-DeepONet and CNN-UNet achieve relatively better agreement but retain noticeable errors. The error distributions highlight the improved accuracy of SpectONet in resolving the simply supported beam response.

\begin{figure*}
    \centering
    \includegraphics[width=0.8\textwidth]{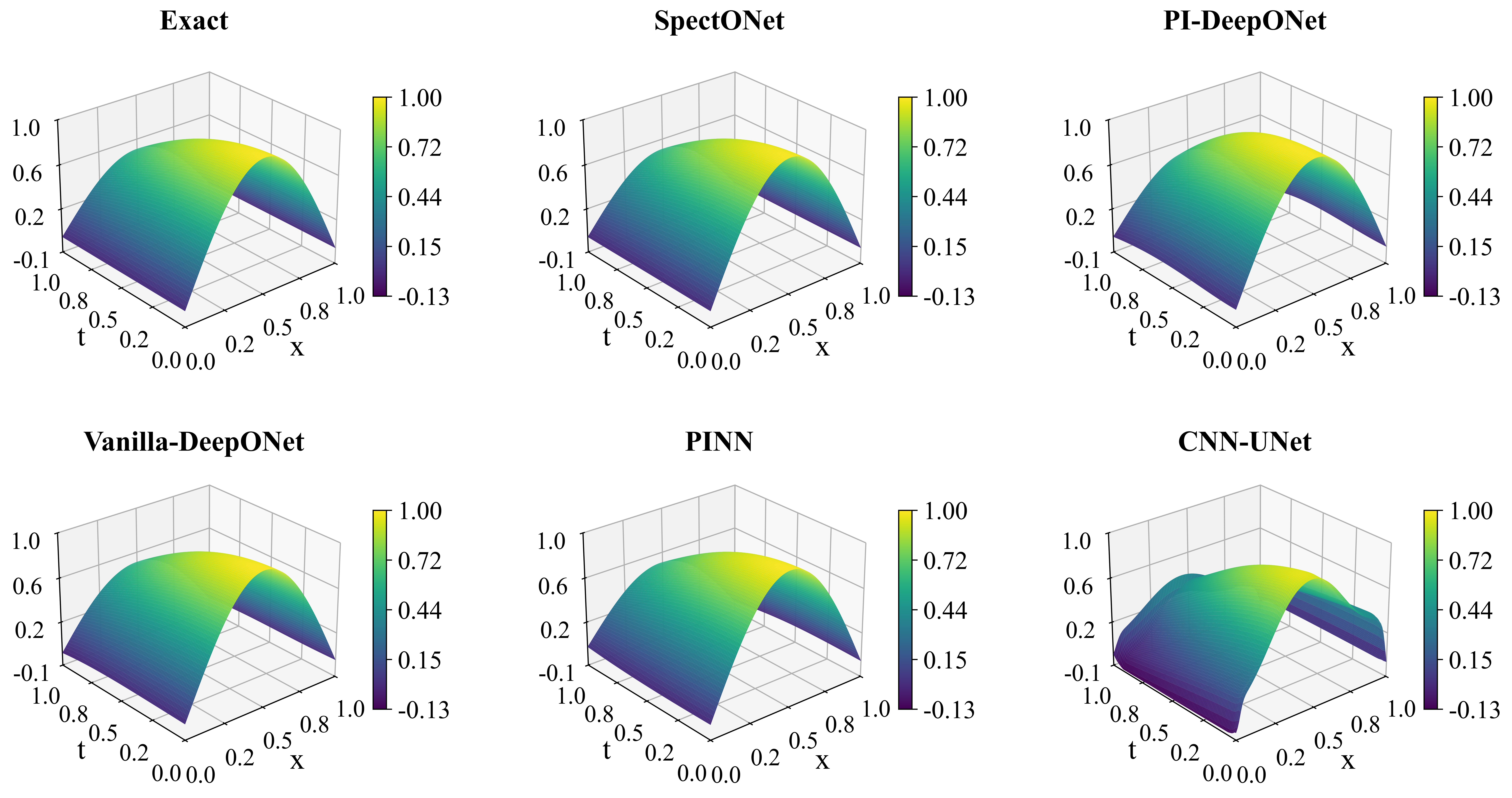}
    \caption{3D comparison of the exact solution and the predicted solutions obtained by the proposed SpectONet with the baseline model for P2 over the spatio-temporal domain $(x,t)\in[0,1]\times[0,1].$}
    \label{3d_solution_comparison_P2}
\end{figure*}

Figure~\ref{3d_solution_comparison_P2} illustrates the 3D displacement surfaces obtained from SpectONet and the baseline models for P2. The SpectONet prediction accurately captures the sinusoidal spatial profile and its temporal evolution, exhibiting close agreement with the analytical solution. In contrast, the baseline models show varying degrees of deviation in their predicted displacement surfaces, reflecting differences in their ability to approximate the beam vibration response.

The quantitative results for P2 are presented in
Table~\ref{tab:problem2_errors}, which compares
the performance of SpectONet and four baseline
models across four independent random seeds
S1-S4. The relative $\mathcal{L}_2$ error,
RMSE, MAE, and P95 error are reported as mean
values with their corresponding standard deviations,
whereas the maximum error is reported as the mean
across the four seeds. SpectONet achieves a mean
relative $\mathcal{L}_2$ error of $1.025e{-3}$,
corresponding to reductions of approximately
86.27\%, 97.47\%, 96.67\%, and 99.20\%
compared with Vanilla-DeepONet, PI-DeepONet,
PINN, and CNN-UNet, respectively. The proposed
model also achieves the lowest mean RMSE, MAE,
P95 error, and maximum error among the
considered models.

\begin{table*} 
\centering
\caption{Performance comparison of the proposed model with the baseline models for P2 (mean $\pm$ standard deviation over four seeds, S1-S4).}
\label{tab:problem2_errors}
\renewcommand{\arraystretch}{1.2}
\setlength{\tabcolsep}{3pt}

\resizebox{\textwidth}{!}{%
\begin{tabular}{lcccccc}
\hline
\textbf{Model} &
\textbf{Rel. $\mathcal{L}_2$ Error} &
\textbf{RMSE} &
\textbf{MAE} &
\textbf{P95 Error} &
\textbf{Max. Error} \\
\hline

\textbf{SpectONet} &
$\mathbf{1.025e{-3}\pm9.193e{-4}}$ &
$\mathbf{6.148e{-4}\pm5.513e{-4}}$ &
$\mathbf{4.912e{-4}\pm4.265e{-4}}$ &
$\mathbf{1.226e{-3}\pm1.211e{-3}}$ &
$\mathbf{2.557e{-3}}$  \\

PI-DeepONet &
$4.059e{-2}\pm1.956e{-2}$ &
$2.434e{-2}\pm1.173e{-2}$ &
$2.056e{-2}\pm1.001e{-2}$ &
$4.112e{-2}\pm1.946e{-2}$ &
$5.473e{-2}$  \\

Vanilla-DeepONet &
$7.468e{-3}\pm3.231e{-3}$ &
$4.478e{-3}\pm1.937e{-3}$ &
$3.396e{-3}\pm1.517e{-3}$ &
$8.190e{-3}\pm3.054e{-3}$ &
$2.879e{-2}$ \\

PINN &
$3.077e{-2}\pm3.392e{-2}$ &
$1.845e{-2}\pm2.035e{-2}$ &
$1.278e{-2}\pm1.358e{-2}$ &
$4.145e{-2}\pm4.918e{-2}$ &
$8.786e{-2}$  \\

CNN-UNet &
$1.285e{-1}\pm6.414e{-2}$ &
$7.705e{-2}\pm3.847e{-2}$ &
$6.574e{-2}\pm3.746e{-2}$ &
$1.367e{-1}\pm5.244e{-2}$ &
$1.975e{-1}$ \\

\hline
\end{tabular}%
}
\end{table*}

\begin{figure*} 
    \centering
    \includegraphics[width=0.95\textwidth]{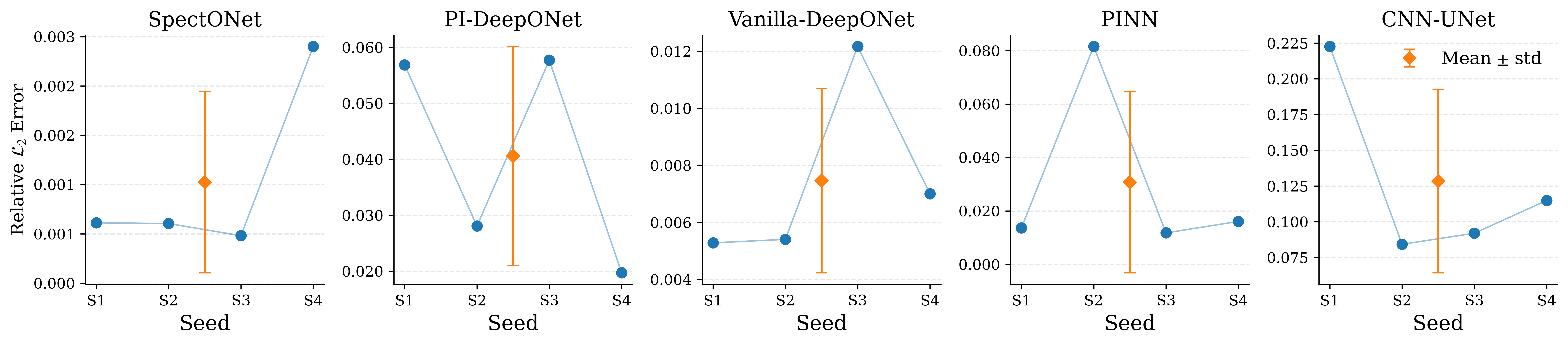}
    \caption{Comparison of the relative $\mathcal{L}_2$ errors of SpectONet and four baseline models across four independent random seeds (S1-S4) for P2.}
    \label{fig:p2_seed_robustness}
\end{figure*}

The effect of random initialization on model
performance is further examined in
Figure~\ref{fig:p2_seed_robustness}, which
illustrates the relative $\mathcal{L}_2$ errors
of SpectONet and four baseline models across
S1-S4. This seed-wise comparison provides
further insight into the variation in predictive
performance under different random initializations
and supports the overall accuracy of SpectONet.

\subsection{Forced Undamped Initially Resting Simply Supported Euler-Bernoulli Beam}
\label{P3}

The third benchmark problem (P3) considers the forced
vibration of an initially resting simply supported EBB,
as illustrated in Figure~\ref{All_beam}.
The governing equation is given by
Eq.~\eqref{OL_undamped}, with an external harmonic
excitation of the form
\begin{equation}
F(x,t)=
f_0\sin\left(\frac{n\pi x}{L}\right)
\sin(\Omega t),
\label{P3_force}
\end{equation}
where $f_0$ denotes the forcing amplitude,
$n$ is the mode number, $L$ is the beam length,
and $\Omega$ is the excitation frequency.

The physical and geometric parameters are
chosen as follows:~\cite{ccayan2022collocation}
\begin{equation}
\begin{gathered}
E=207\times10^9~\mathrm{Pa},\quad
I=14.06\times10^{-6}~\mathrm{m}^4,\\
\rho=76500~\mathrm{kg/m^3},\quad
A=0.0075~\mathrm{m}^2,\\
L=1~\mathrm{m},\quad f_0=10~\mathrm{N/m},\\
\Omega=1~\mathrm{rad/s},\quad
n=1,\quad T=1~\mathrm{s}.
\end{gathered}
\label{P3_parameters}
\end{equation}
The homogeneous simply supported BCs and the corresponding ICs are prescribed in Eqs.~\eqref{P2_BC} and \eqref{P3_IC}, respectively.

\noindent
For the prescribed harmonic excitation,
the analytical solution is given
by~\cite{ccayan2022collocation}
\begin{equation}
\begin{aligned}
u(x,t)
={}&
\frac{f_0L^4}
{EI(n\pi)^4
\left[1-\left(\frac{\Omega}{\omega_n}\right)^2\right]}
\sin\left(\frac{n\pi x}{L}\right)
\\
&\times
\left[
\sin(\Omega t)
-
\frac{\Omega}{\omega_n}
\sin(\omega_n t)
\right],
\end{aligned}
\label{P3_exact}
\end{equation}
where the natural angular frequency is defined as:
\begin{equation}
\omega_n=n^2\pi^2
\sqrt{\frac{EI}{\rho A L^4}},
\qquad \Omega\neq\omega_n.
\end{equation}
For the analytical solution in
Eq.~\eqref{P3_exact}, the corresponding
spatial input is
$g(x)=f_0\sin(n\pi x/L)$.

\noindent
The PDE residual loss for P3 is defined as:
\begin{equation}
\begin{aligned}
\mathcal{L}_{\mathrm{PDE}}
={}&
\frac{1}{N_fN_r}
\sum_{i=1}^{N_f}
\sum_{j=1}^{N_r}
\Bigg|
\rho A
\frac{\partial^2
\widehat{\mathcal{G}}_{\theta}(g^{(i)})
(x_j^r,t_j^r)}
{\partial t^2}
\\
&+
EI
\frac{\partial^4
\widehat{\mathcal{G}}_{\theta}(g^{(i)})
(x_j^r,t_j^r)}
{\partial x^4}
-
g^{(i)}(x_j^r)\sin(\Omega t_j^r)
\Bigg|^2.
\end{aligned}
\label{P3_PDE_loss}
\end{equation}

\noindent
The IC loss is defined as:
\begin{equation}
\begin{aligned}
\mathcal{L}_{\mathrm{IC}}
={}&
\frac{1}{N_fN_0}
\sum_{i=1}^{N_f}
\sum_{j=1}^{N_0}
\Bigg[
\left|
\widehat{\mathcal{G}}_{\theta}(g^{(i)})
(x_j^0,0)
\right|^2
\\
&+
\left|
\frac{\partial
\widehat{\mathcal{G}}_{\theta}(g^{(i)})
(x_j^0,0)}
{\partial t}
\right|^2
\Bigg].
\end{aligned}
\label{P3_IC_loss}
\end{equation}
The boundary loss is introduced to enforce the homogeneous simply supported conditions and is given by
\begin{equation}
\begin{aligned}
\mathcal{L}_{\mathrm{BC}}
={}&
\frac{1}{N_fN_b}
\sum_{i=1}^{N_f}
\sum_{j=1}^{N_b}
\Bigg[
\left|
\widehat{\mathcal{G}}_{\theta}(g^{(i)})(0,t_j^b)
\right|^2
+
\left|
\widehat{\mathcal{G}}_{\theta}(g^{(i)})(L,t_j^b)
\right|^2
\\
&+
\left|
\frac{\partial^2
\widehat{\mathcal{G}}_{\theta}(g^{(i)})(0,t_j^b)}
{\partial x^2}
\right|^2
+
\left|
\frac{\partial^2
\widehat{\mathcal{G}}_{\theta}(g^{(i)})(L,t_j^b)}
{\partial x^2}
\right|^2
\Bigg].
\end{aligned}
\label{P3_BC_loss}
\end{equation}

\noindent
The total loss for P3 follows Eq.~\eqref{total_loss}.
The network architecture and problem-specific training
hyperparameters adopted for P3 are summarized in
Table~\ref{SpectONet_parameters}. To enforce the
governing equation together with the prescribed ICs
and BCs, 1024 interior collocation points, 100 initial
points, and 100 boundary points are randomly sampled
from the spatio-temporal domain
$(x,t)\in[0,1]\times[0,1]$.

\begin{figure*} 
\centering
\setlength{\tabcolsep}{4pt}
\renewcommand{\arraystretch}{1.5}

\fbox{
\begin{tabular}{c|c}
\textbf{Predicted Solutions} &
\textbf{Absolute Error} \\ \hline

\includegraphics[width=0.47\textwidth]{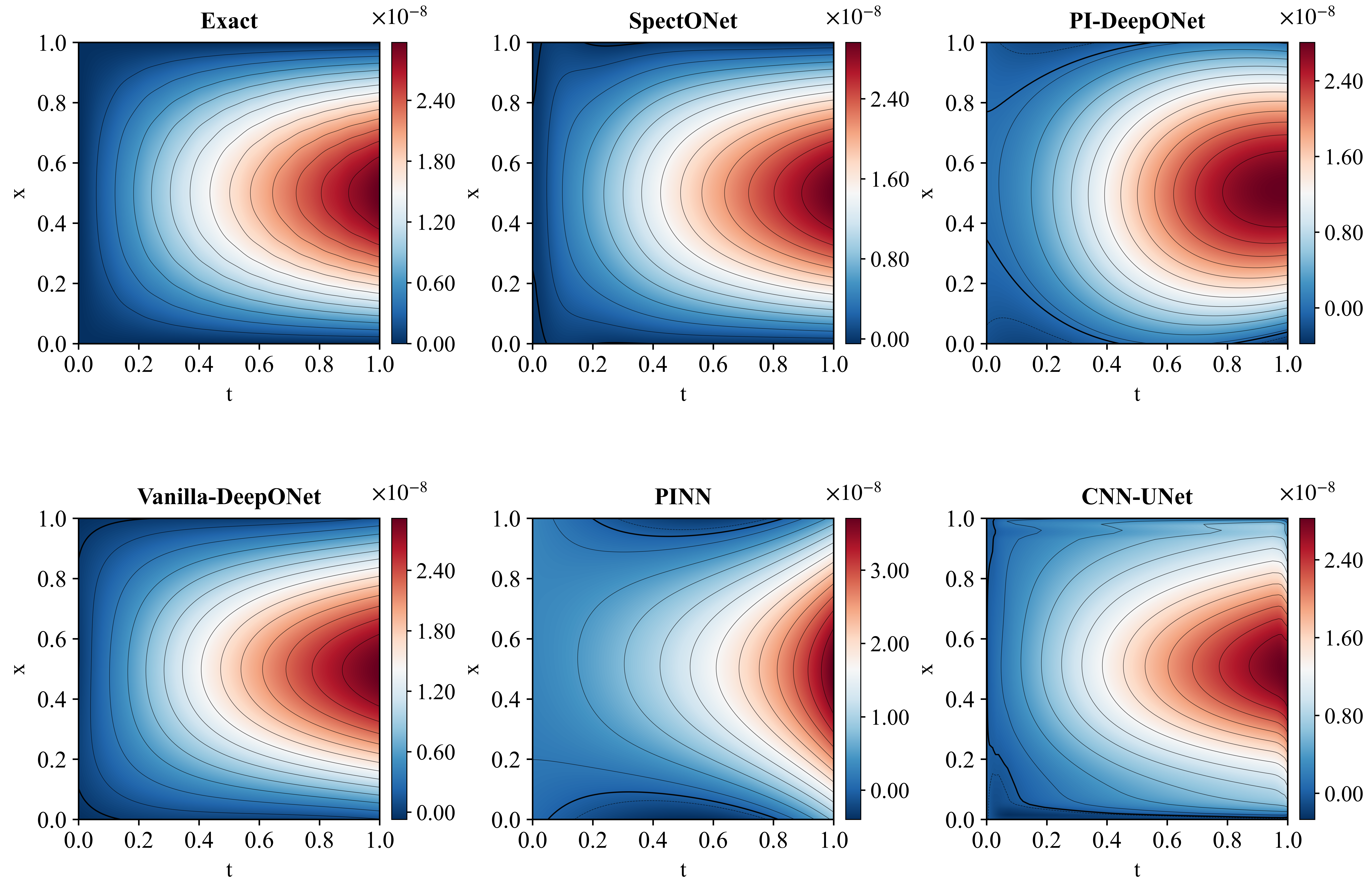} &
\includegraphics[width=0.47\textwidth]{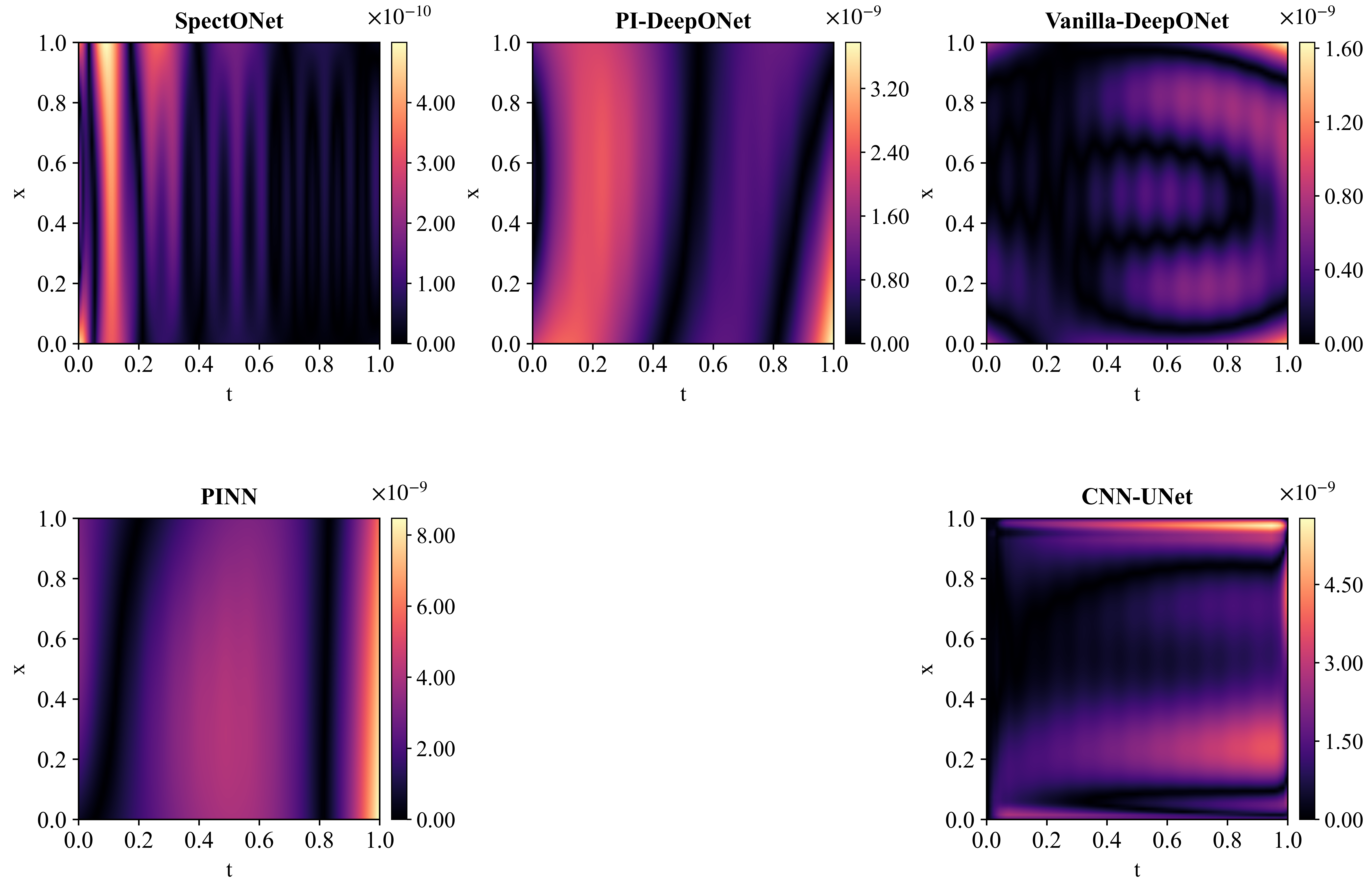} \\

\end{tabular}
}

\caption{
Qualitative comparison of the proposed SpectONet with the baseline models for P3. The left column presents the 2D predicted solution fields, whereas the right column shows the corresponding point-wise absolute error distributions over the spatio-temporal domain $(x,t)\in[0,1]\times[0,1].$
}
\label{fig:P3_box}
\end{figure*}

The displacement response of the initially resting beam is examined through the 2D solution fields and absolute error distributions presented in Figure~\ref{fig:P3_box}. SpectONet captures the forced vibration response with close agreement to the analytical reference, while the competing models exhibit varying levels of prediction error. The error maps provide a spatial and temporal assessment of the differences between the numerical predictions and the exact displacement field.

\begin{figure*} 
    \centering
    \includegraphics[width=0.8\textwidth]{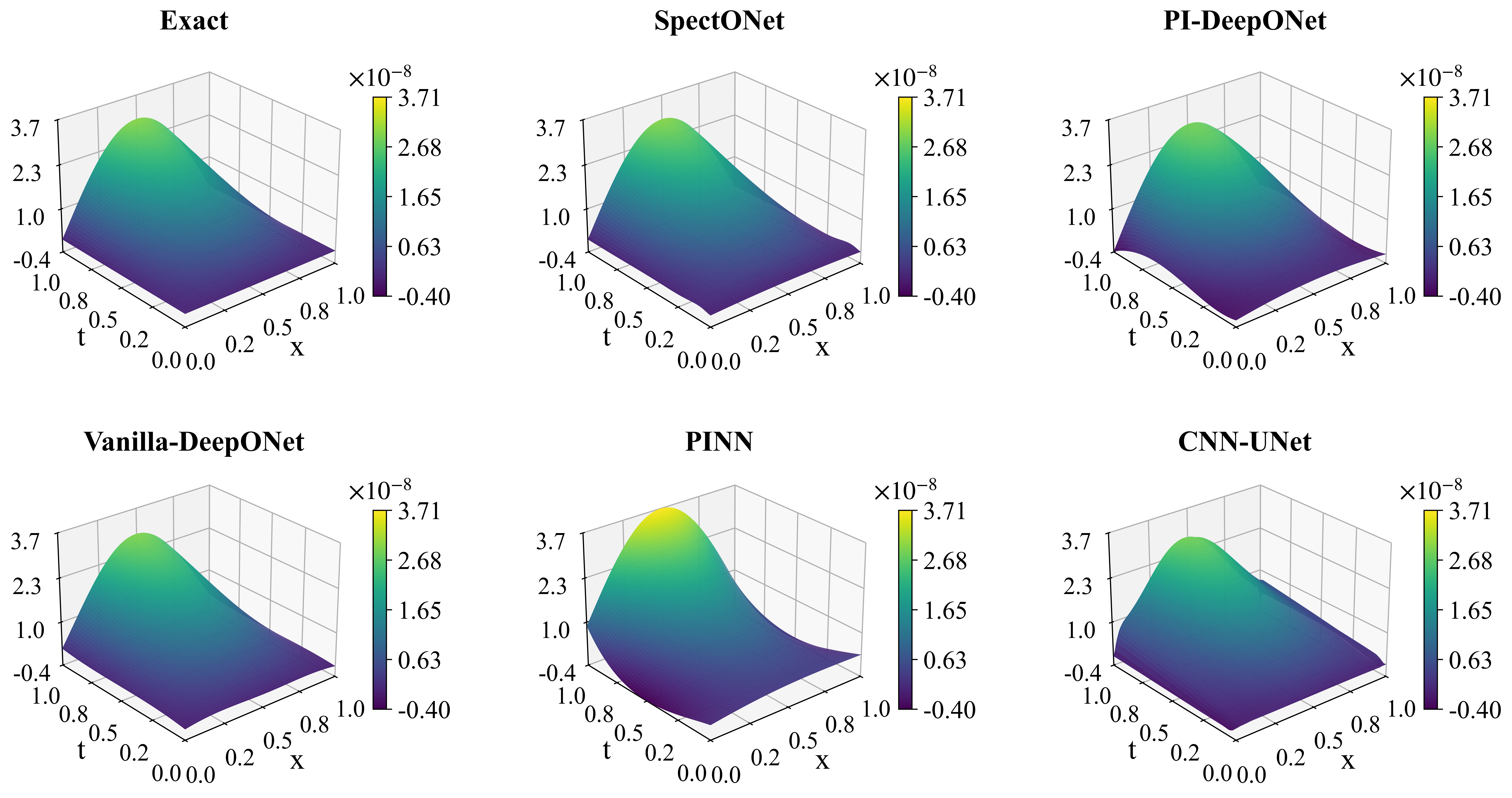}
    \caption{3D comparison of the exact solution and the predicted solutions obtained by the proposed spectONet with the baseline model for P3 over the spatio-temporal domain $(x,t)\in[0,1]\times[0,1].$}
    \label{3d_solution_comparison_P3}
\end{figure*}

Figure~\ref{3d_solution_comparison_P3} displays the 3D displacement surfaces for the considered models. The surface obtained using SpectONet follows the reference vibration profile, reproducing the displacement variations induced by the harmonic excitation. Differences between the predicted surfaces are also evident for the baseline approaches.

Unlike P1, which considers nonhomogeneous BCs, P3 involves homogeneous simply supported
BCs, similar to P2. The quantitative
results presented in Table~\ref{tab:problem3_errors}
show that SpectONet achieves a mean relative
$\mathcal{L}_2$ error of $1.647e{-2}$, corresponding
to reductions of 56.54\%, 88.43\%,
87.65\%, and 86.50\% compared with
Vanilla-DeepONet, PI-DeepONet, PINN, and CNN-UNet,
respectively. Furthermore, SpectONet attains
the lowest mean RMSE, MAE, P95 error, and
maximum error among the considered models.

\begin{table*} 
\centering
\caption{Performance comparison of the proposed model with the baseline models for P3 (mean $\pm$ standard deviation over four seeds, S1-S4).}
\label{tab:problem3_errors}
\renewcommand{\arraystretch}{1.2}
\setlength{\tabcolsep}{3pt}
\resizebox{\textwidth}{!}{%
\begin{tabular}{lcccccc}
\hline
\textbf{Model} &
\textbf{Rel. $\mathcal{L}_2$ Error} &
\textbf{RMSE} &
\textbf{MAE} &
\textbf{P95 Error} &
\textbf{Max. Error}\\
\hline

\textbf{SpectONet} &
$\mathbf{1.647e{-2}\pm8.141e{-3}}$ &
$\mathbf{2.137e{-10}\pm1.057e{-10}}$ &
$\mathbf{1.414e{-10}\pm6.437e{-11}}$ &
$\mathbf{4.813e{-10}\pm2.278e{-10}}$ &
$\mathbf{1.005e{-9}}$  \\

PI-DeepONet &
$1.424e{-1}\pm6.773e{-2}$ &
$1.849e{-9}\pm8.791e{-10}$ &
$1.595e{-9}\pm7.864e{-10}$ &
$3.136e{-9}\pm1.448e{-9}$ &
$5.214e{-9}$ \\

Vanilla-DeepONet &
$3.790e{-2}\pm3.765e{-2}$ &
$4.919e{-10}\pm4.886e{-10}$ &
$3.706e{-10}\pm3.709e{-10}$ &
$9.260e{-10}\pm9.323e{-10}$ &
$2.739e{-9}$ \\

PINN &
$1.334e{-1}\pm6.933e{-2}$ &
$1.731e{-9}\pm8.999e{-10}$ &
$1.456e{-9}\pm8.091e{-10}$ &
$2.924e{-9}\pm1.430e{-9}$ &
$4.247e{-9}$ \\

CNN-UNet &
$1.220e{-1}\pm2.974e{-2}$ &
$1.584e{-9}\pm3.861e{-10}$ &
$1.243e{-9}\pm3.414e{-10}$ &
$3.090e{-9}\pm7.330e{-10}$ &
$5.821e{-9}$ \\

\hline
\end{tabular}
}
\end{table*}

\begin{figure*} 
    \centering
    \includegraphics[width=0.95\textwidth]{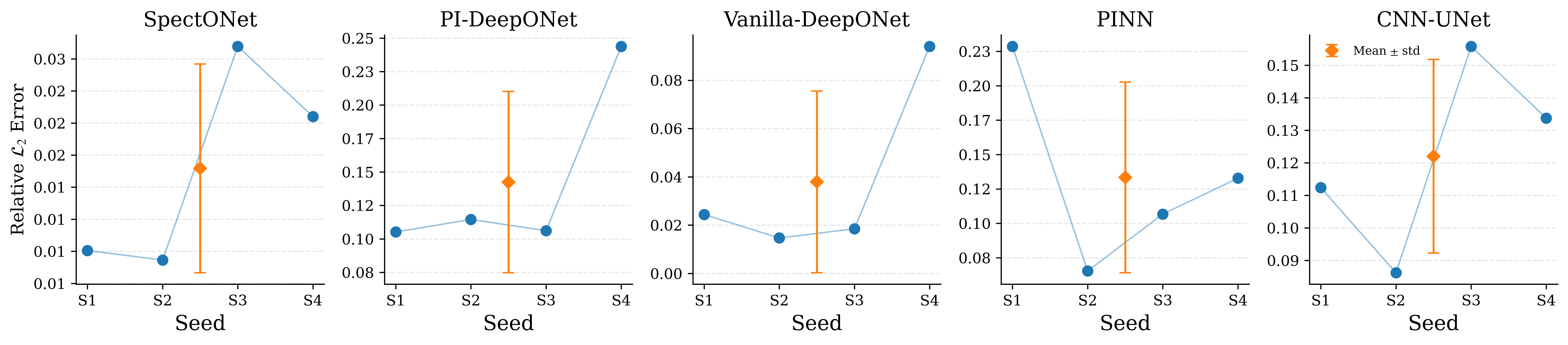}
   
\caption{Comparison of the relative $\mathcal{L}_2$ errors of SpectONet and four baseline models across four independent random seeds (S1-S4) for P3.}

    \label{fig:p3_seed_robustness}
\end{figure*}

Figure~\ref{fig:p3_seed_robustness} presents
the relative $\mathcal{L}_2$ errors of SpectONet
and four baseline models across S1-S4.
The seed-wise results provide further insight
into the variation in predictive accuracy under
different network initializations and support
the overall predictive performance of SpectONet.

\section{Results and Discussion}
\label{sec_6}

This section evaluates the performance of SpectONet
on three EBB vibration problems. Furthermore,
an ablation study on P1 investigates the influence
of loss weights, collocation-point density, and
individual physics-informed loss components on the
predictive accuracy of SpectONet.

\subsection{Spatial Displacement Profiles}

To examine the spatial accuracy of the predicted displacement fields, the solutions obtained using SpectONet and the baseline models are compared with the analytical reference solutions at selected time instants. The corresponding spatial profiles
and quantitative error comparisons are presented separately for P1, P2, and P3.

For P1, Figure~\ref{fig:P1_time_profiles} compares the spatial displacement profiles predicted by SpectONet and the four baseline models. At $t=0.45$, SpectONet exhibits remarkable accuracy, achieving a relative $\mathcal{L}_2$ error approximately $58.51$ times lower than CNN-UNet, the most accurate baseline at this time instant. At $t=0.70$, CNN-UNet continues to outperform the other baseline models; however, its relative error remains approximately $12.11$ times higher than that of SpectONet. Interestingly, at $t=1.00$, Vanilla-DeepONet surpasses CNN-UNet and becomes the best-performing baseline, although SpectONet still achieves approximately $8.77$ times lower relative error. While the relative error of SpectONet increases with time, its predictions consistently exhibit the closest agreement with the analytical displacement profiles. The quantitative results presented in Table~\ref{tab:P1_selected_time_error_metrics} further establish its superior accuracy across the selected time instants.

\begin{figure*} 
    \centering
    \includegraphics[width=0.95\textwidth]{Different_t_P1.png}
    
\caption{Comparison of the exact and predicted solutions of SpectONet and four baseline models at three different time instants for P1.}

    \label{fig:P1_time_profiles}
\end{figure*}

\begin{table}[!t]
\centering
\caption{Comparison of error metrics for different models at selected 
time instants over the spatial domain $x\in[0,1]$ for P1.}
\label{tab:P1_selected_time_error_metrics}
\renewcommand{\arraystretch}{1.3}

\resizebox{\columnwidth}{!}{
\begin{tabular}{lccccc}
\hline

\textbf{Model} &
\textbf{Time ($t$)} &
\textbf{Rel. $\mathcal{L}_2$ Error} &
\textbf{RMSE} &
\textbf{MAE} &
\textbf{Max. Error} \\
\hline

\multirow{3}{*}{SpectONet}
& 0.45 & 6.011e-03 & 2.145e-04 & 1.940e-04 & 3.099e-04 \\
& 0.70 & 2.593e-02 & 7.864e-04 & 7.270e-04 & 1.075e-03 \\
& 1.00 & 5.828e-02 & 1.248e-03 & 1.090e-03 & 1.901e-03 \\
\hline

\multirow{3}{*}{PI-DeepONet}
& 0.45 & 6.250e+00 & 2.231e-01 & 1.818e-01 & 5.283e-01 \\
& 0.70 & 1.452e+01 & 4.403e-01 & 3.918e-01 & 6.238e-01 \\
& 1.00 & 6.027e+01 & 1.291e+00 & 1.270e+00 & 1.516e+00 \\
\hline

\multirow{3}{*}{Vanilla-DeepONet}
& 0.45 & 4.984e-01 & 1.779e-02 & 1.595e-02 & 2.736e-02 \\
& 0.70 & 4.948e-01 & 1.500e-02 & 1.344e-02 & 2.298e-02 \\
& 1.00 & 5.110e-01 & 1.094e-02 & 9.464e-03 & 1.901e-02 \\
\hline

\multirow{3}{*}{PINN}
& 0.45 & 2.710e+01 & 9.675e-01 & 9.672e-01 & 1.044e+00 \\
& 0.70 & 1.443e+01 & 4.376e-01 & 4.358e-01 & 5.601e-01 \\
& 1.00 & 5.336e+01 & 1.143e+00 & 1.142e+00 & 1.195e+00 \\
\hline

\multirow{3}{*}{CNN-UNet}
& 0.45 & 3.517e-01 & 1.255e-02 & 1.090e-02 & 2.345e-02 \\
& 0.70 & 3.140e-01 & 9.522e-03 & 8.054e-03 & 2.495e-02 \\
& 1.00 & 7.245e-01 & 1.552e-02 & 1.315e-02 & 2.462e-02 \\
\hline

\end{tabular}
}
\end{table}

For P2, Figure~\ref{fig:P2_time_profiles} illustrates
the spatial displacement profiles predicted by SpectONet
and the four baseline models. At $t=0.45$, SpectONet
achieves a relative $\mathcal{L}_2$ error approximately
$4.88$ times smaller than that of Vanilla-DeepONet,
the most accurate baseline. At $t=0.70$, SpectONet
attains its lowest relative error of
$3.0e-04$, which is approximately
$15.16$ times smaller than that of Vanilla-DeepONet.
Although the relative error of SpectONet increases
at $t=1.00$, the corresponding error of
Vanilla-DeepONet remains approximately $10.26$
times higher than that of SpectONet. Notably,
Vanilla-DeepONet consistently outperforms the
other baseline models, whereas CNN-UNet exhibits
the largest relative errors at all three time
instants. These observations indicate that
SpectONet more accurately captures the sinusoidal
displacement response of the simply supported beam,
as further supported by the error metrics reported
in Table~\ref{tab:P2_selected_time_error_metrics}.

\begin{figure*}[t]
    \centering
    \includegraphics[width=0.95\textwidth]{Different_t_P2.png}
    \caption{Comparison of the exact and predicted solutions of SpectONet and four baseline models at three different time instants for P2.}
    \label{fig:P2_time_profiles}
\end{figure*}

\begin{table} 
\centering
\caption{Comparison of error metrics for different models at selected 
time instants over the spatial domain $x\in[0,1]$ for P2.}
\label{tab:P2_selected_time_error_metrics}
\renewcommand{\arraystretch}{1.3}

\resizebox{\columnwidth}{!}{
\begin{tabular}{lccccc}
\hline

\textbf{Model} &
\textbf{Time ($t$)} &
\textbf{Rel. $\mathcal{L}_2$ Error} &
\textbf{RMSE} &
\textbf{MAE} &
\textbf{Max. Error} \\
\hline

\multirow{3}{*}{SpectONet}
& 0.45 & 8.381e-04 & 5.310e-04 & 4.730e-04 & 7.190e-04 \\
& 0.70 & 3.009e-04 & 1.619e-04 & 1.368e-04 & 3.121e-04 \\
& 1.00 & 1.883e-03 & 7.160e-04 & 4.839e-04 & 1.914e-03 \\
\hline

\multirow{3}{*}{PI-DeepONet}
& 0.45 & 7.335e-02 & 4.647e-02 & 4.637e-02 & 4.972e-02 \\
& 0.70 & 8.554e-02 & 4.603e-02 & 4.593e-02 & 4.950e-02 \\
& 1.00 & 6.098e-02 & 2.318e-02 & 1.883e-02 & 4.431e-02 \\
\hline

\multirow{3}{*}{Vanilla-DeepONet}
& 0.45 & 4.090e-03 & 2.591e-03 & 2.212e-03 & 7.887e-03 \\
& 0.70 & 4.563e-03 & 2.455e-03 & 2.129e-03 & 4.017e-03 \\
& 1.00 & 1.931e-02 & 7.344e-03 & 4.585e-03 & 2.543e-02 \\
\hline

\multirow{3}{*}{PINN}
& 0.45 & 7.744e-03 & 4.906e-03 & 4.033e-03 & 1.110e-02 \\
& 0.70 & 5.539e-03 & 2.981e-03 & 2.758e-03 & 4.235e-03 \\
& 1.00 & 6.030e-02 & 2.292e-02 & 1.946e-02 & 4.738e-02 \\
\hline

\multirow{3}{*}{CNN-UNet}
& 0.45 & 1.975e-01 & 1.251e-01 & 1.163e-01 & 2.127e-01 \\
& 0.70 & 2.811e-01 & 1.513e-01 & 1.434e-01 & 2.125e-01 \\
& 1.00 & 2.268e-01 & 8.623e-02 & 7.379e-02 & 1.431e-01 \\
\hline

\end{tabular}
}
\end{table}

For P3, Figure~\ref{fig:P3_time_profiles} illustrates
the spatial displacement profiles of the initially
resting simply supported beam under harmonic excitation,
as predicted by SpectONet and the four baseline models.
At $t=0.45$, SpectONet achieves a relative
$\mathcal{L}_2$ error of $9.0e-03$,
approximately $2.40$ times smaller than that of
Vanilla-DeepONet, the most accurate baseline.
At $t=0.70$, the relative error of SpectONet
decreases to $2.2e-03$, which is
approximately $10.61$ times smaller than that of
Vanilla-DeepONet. Although the relative error of
SpectONet increases slightly at $t=1.00$,
the corresponding error of Vanilla-DeepONet
remains approximately $11.05$ times higher
than that of SpectONet. Among the baseline models,
PINN exhibits the largest relative errors at all
three time instants, whereas Vanilla-DeepONet
consistently provides the closest approximation
to the analytical response. Furthermore, the RMSE
values obtained using SpectONet range from
$3.6e-11$ to $9.7e-11$,
indicating small absolute discrepancies in the
predicted displacement field. The detailed error
comparisons are presented in
Table~\ref{tab:P3_selected_time_error_metrics}.

\begin{figure*} 
    \centering
    \includegraphics[width=0.95\textwidth]{Different_t_P3.png}
    \caption{Comparison of the exact and predicted solutions of SpectONet and four baseline models at three different time instants for P3.}
    \label{fig:P3_time_profiles}
\end{figure*}

\begin{table} 
\centering
\caption{Comparison of error metrics for different models at selected 
time instants over the spatial domain $x\in[0,1]$ for P3.}
\label{tab:P3_selected_time_error_metrics}
\renewcommand{\arraystretch}{1.3}

\resizebox{\columnwidth}{!}{
\begin{tabular}{lccccc}
\hline

\textbf{Model} &
\textbf{Time ($t$)} &
\textbf{Rel. $\mathcal{L}_2$ Error} &
\textbf{RMSE} &
\textbf{MAE} &
\textbf{Max. Error} \\
\hline

\multirow{3}{*}{SpectONet}
& 0.45 & 9.018e-03 & 9.709e-11 & 9.480e-11 & 1.219e-10 \\
& 0.70 & 2.276e-03 & 3.632e-11 & 3.277e-11 & 5.145e-11 \\
& 1.00 & 2.721e-03 & 5.690e-11 & 4.953e-11 & 8.448e-11 \\
\hline

\multirow{3}{*}{PI-DeepONet}
& 0.45 & 6.242e-02 & 6.720e-10 & 6.106e-10 & 8.821e-10 \\
& 0.70 & 6.226e-02 & 9.935e-10 & 9.930e-10 & 1.028e-09 \\
& 1.00 & 9.906e-02 & 2.071e-09 & 1.725e-09 & 3.778e-09 \\
\hline

\multirow{3}{*}{Vanilla-DeepONet}
& 0.45 & 2.165e-02 & 2.330e-10 & 2.061e-10 & 4.347e-10 \\
& 0.70 & 2.415e-02 & 3.854e-10 & 3.382e-10 & 6.518e-10 \\
& 1.00 & 3.007e-02 & 6.288e-10 & 5.509e-10 & 1.632e-09 \\
\hline

\multirow{3}{*}{PINN}
& 0.45 & 3.441e-01 & 3.704e-09 & 3.681e-09 & 4.066e-09 \\
& 0.70 & 1.612e-01 & 2.572e-09 & 2.571e-09 & 2.640e-09 \\
& 1.00 & 3.502e-01 & 7.322e-09 & 7.285e-09 & 8.472e-09 \\
\hline

\multirow{3}{*}{CNN-UNet}
& 0.45 & 1.154e-01 & 1.243e-09 & 1.007e-09 & 3.471e-09 \\
& 0.70 & 1.042e-01 & 1.662e-09 & 1.387e-09 & 4.610e-09 \\
& 1.00 & 1.067e-01 & 2.230e-09 & 1.964e-09 & 3.995e-09 \\
\hline

\end{tabular}
}
\end{table}

The time-specific comparisons reveal distinct error characteristics across the three benchmark problems. SpectONet consistently achieves the lowest relative $\mathcal{L}_2$ errors at the selected time instants, while the baseline models exhibit larger discrepancies depending on the loading conditions and displacement profiles. These findings extend the earlier global error analysis by demonstrating the predictive accuracy of SpectONet at individual time instants.

\subsection{Ablation Study}

To further investigate the robustness and sensitivity of the proposed SpectONet, an ablation study is conducted on P1 only. Three important aspects that directly influence the training process are investigated: (i) loss-weight configurations; (ii) number of collocation points; and (iii) sensitivity of loss components.

\subsubsection{Loss-Weight Configuration}

The predictive performance of physics-informed neural operator models can be sensitive to the relative weighting of the governing PDE residual and the associated BC and IC losses. To investigate this sensitivity, several loss-weight configurations were evaluated while keeping the remaining training settings fixed. As described in Eq.~\eqref{total_loss}, the BC and IC losses are further decomposed into their individual constraints. The seven loss weights are ordered as $(\lambda_r,\lambda_{b_1},\lambda_{b_2},\lambda_{b_3},\lambda_{b_4},\lambda_{i_1},\lambda_{i_2})$, corresponding to the PDE residual, four individual BC constraints, and two individual IC constraints, respectively. The corresponding prediction errors are summarized in Table~\ref{tab:loss}, where the highlighted row denotes the loss-weight configuration adopted throughout this study.

\begin{table*} 
\centering
\caption{Influence of different loss-weight configurations on the predictive performance of the proposed model for P1.}
\label{tab:loss}
\renewcommand{\arraystretch}{1.5}
\resizebox{\textwidth}{!}{
\begin{tabular}{lccccccccc}
\hline
Configuration &
Seed &
$N_r$ &
$N_b$ &
$N_i$ &
Loss Weights &
Relative $\mathcal{L}_2$ &
RMSE &
MAE &
Max Error \\
\hline

Balanced
&1234
&1024
&128
&128
&$(1,1,1,1,1,1,1)$
&$4.679e-02$
&$1.581e-03$
&$1.302e-03$
&$5.392e-03$ \\

BC/IC-heavy
&1234
&1024
&128
&128
&$(1,20,20,20,20,20,20)$
&$2.977e-02$
&$1.006e-03$
&$7.950e-04$
&$2.834e-03$ \\

PDE-heavy
&1234
&1024
&128
&128
&$(10,10,10,10,10,10,10)$
&$1.873e-02$
&$6.328e-04$
&$4.681e-04$
&$4.428e-03$ \\

Reduced PDE Weight
&1234
&1024
&128
&128
&$(0.1,10,10,10,10,10,10)$
&{$1.589e-02$}
&{$5.332e-04$}
&{$3.901e-04$}
&$3.420e-03$ \\

\rowcolor{gray!15}
\textbf{Default}
&\textbf{1234}
&\textbf{1024}
&\textbf{128}
&\textbf{128}
&$\mathbf{(1,10,10,10,10,10,10)}$
&$\mathbf{1.550e-02}$
&$\mathbf{5.240e-04}$
&$\mathbf{4.161e-04}$
&$\mathbf{1.901e-03}$ \\

\hline
\end{tabular}}
\end{table*}

The predictive performance of the proposed SpectONet is sensitive to the tested loss-weight configurations. The adopted configuration achieves the lowest relative $\mathcal{L}_2$ error, RMSE, and maximum error among the configurations considered. Although the reduced PDE-weight configuration yields a slightly lower MAE, the adopted configuration provides favorable overall performance across the evaluation metrics. The remaining configurations, including balanced, BC/IC-heavy, and uniformly scaled, generally exhibit higher prediction errors. These results indicate that SpectONet is sensitive to the tested loss-weight configurations for P1 under the considered random seed.

\subsubsection{Collocation-Point Density}

The number of collocation points determines how effectively the governing PDE and the associated IC and BC are enforced during training. To investigate the influence of the collocation-point density, several sampling configurations were evaluated while keeping others fixed.

\begin{table*} 
\centering
\caption{Influence of different collocation-point densities on the predictive performance of the proposed model for P1.}
\label{tab:collocation}
\renewcommand{\arraystretch}{1.5}
\resizebox{\textwidth}{!}{
\begin{tabular}{lccccccccc}
\hline
Configuration &
Seed &
$N_r$ &
$N_b$ &
$N_i$ &
Loss Weights &
Relative $\mathcal{L}_2$ &
RMSE &
MAE &
Max Error \\
\hline

Small
&1234
&256
&64
&64
&$(1,10,10,10,10,10,10)$
&{$8.411e-02$}
&{$2.842e-03$}
&{$2.295e-03$}
&{$1.182e-02$} \\

Medium
&1234
&512
&128
&128
&$(1,10,10,10,10,10,10)$
&$2.933e-02$
&$9.910e-04$
&$7.968e-04$
&$3.379e-03$ \\

\rowcolor{gray!15}
\textbf{Default}
&\textbf{1234}
&\textbf{1024}
&\textbf{128}
&\textbf{128}
&$\mathbf{(1,10,10,10,10,10,10)}$
&$\mathbf{1.550e-02}$
&$\mathbf{5.240e-04}$
&$\mathbf{4.161e-04}$
&$\mathbf{1.901e-03}$ \\

Large
&1234
&2048
&256
&256
&$(1,10,10,10,10,10,10)$
&{$7.225e-03$}
&{$2.374e-04$}
&{$1.869e-04$}
&{$6.934e-04$} \\

\hline
\end{tabular}}
\end{table*}

Table~\ref{tab:collocation} presents the influence of
different collocation-point densities on the predictive
performance of the proposed SpectONet. Increasing the
number of collocation points consistently reduces the
prediction errors across all evaluation metrics,
demonstrating a monotonic improvement in predictive
accuracy. The Large configuration achieves the lowest
relative $\mathcal{L}_2$ error, RMSE, MAE, and maximum
error, with substantial improvements over the default
configuration. However, increasing the collocation-point
density also increases the computational requirements
during training. The adopted default configuration,
highlighted in the table, provides a practical balance
between prediction accuracy and computational cost
and is therefore retained for subsequent experiments.

\subsubsection{Sensitivity of Loss Components}
To understand the contribution of each loss component in the proposed SpectONet framework, the individual roles of the physics-residual loss $\mathcal{L}_{\mathrm{PDE}}$, IC loss $\mathcal{L}_{\mathrm{IC}}$, and BC loss $\mathcal{L}_{\mathrm{BC}}$ are investigated by selectively removing each term while keeping all other training settings unchanged. Excluding the physics-residual loss converts the model into a predominantly data-driven framework, which may weaken its physical consistency and generalization capability. Similarly, removing the IC loss can lead to an incorrectly initialized solution whose error propagates over time, while excluding the BC loss may produce noticeable violations near the spatial boundaries. The results reported in Table~\ref{tab:loss_component_results} demonstrate that the best predictive performance is achieved when all loss components are incorporated jointly with appropriate weights.

For a fair comparison, the random seed, network architecture, loss weights of the retained components, collocation-point distribution, optimizer settings, and number of training iterations are kept unchanged across all configurations. This systematic comparison makes it possible to identify the individual contribution of each loss component and determine whether the physics residual, IC, and BC constraints act independently or cooperatively in improving the accuracy and physical consistency of the SpectONet predictions.













\begin{table}[t]
\centering
\caption{Sensitivity of different loss components in SpectONet for P1.}
\label{tab:loss_component_results}

\renewcommand{\arraystretch}{1.2}

\resizebox{\columnwidth}{!}{%
\begin{tabular}{lcccc}
\hline
\textbf{Configuration}
&
\textbf{Physics Loss}
&
\textbf{IC Loss}
&
\textbf{BC Loss}
&
\textbf{Relative $\mathcal{L}_2$ Error}
\\
\hline

\rowcolor{gray!15}
\textbf{Default}
&
$\checkmark$
&
$\checkmark$
&
$\checkmark$
&
$\mathbf{1.550e-02}$
\\

Physics only
&
$\checkmark$
&
$\times$
&
$\times$
&
$1.243e-01$
\\

Protocol 1
&
$\times$
&
$\checkmark$
&
$\times$
&
$7.800e+01$
\\

Protocol 2
&
$\times$
&
$\times$
&
$\checkmark$
&
$8.980e+02$
\\

Protocol 3
&
$\checkmark$
&
$\checkmark$
&
$\times$
&
$3.634e-02$
\\

Protocol 4
&
$\checkmark$
&
$\times$
&
$\checkmark$
&
$5.359e-02$
\\

Protocol 5
&
$\times$
&
$\checkmark$
&
$\checkmark$
&
$2.634e+01$
\\

Protocol 6
&
$\times$
&
$\times$
&
$\times$
&
$6.634e+03$
\\

\hline
\end{tabular}%
}

\end{table}


\subsubsection{Summary}

The ablation study highlights the influence of loss-weight selection, collocation-point density, and individual physics-informed constraints on the predictive performance of SpectONet for P1. Among the investigated weighting strategies, the adopted configuration achieves the lowest prediction errors, emphasizing the importance of appropriately balancing the governing PDE residual with the IC and BC constraints. The collocation-point analysis reveals a non-monotonic relationship between sampling density and prediction accuracy, with only marginal improvements observed when the number of collocation points is increased beyond the default configuration. Furthermore, the loss-component sensitivity analysis demonstrates that jointly enforcing the PDE, IC, and BC losses yields substantially better accuracy than omitting any of these constraints. Collectively, these findings support the effectiveness of the selected training configuration and underscore the complementary roles of the physics-informed loss components in accurately approximating the beam displacement response.

\section{Conclusion}
\label{sec_8}
\justifying

Accurate prediction of structural vibration responses remains a challenging task in scientific machine learning. To address this challenge, the present study proposed SpectONet, a physics-guided spectral deep operator network. The effectiveness of SpectONet was evaluated on three EBB vibration problems involving different forcing functions, ICs, and BCs. Its predictive performance was systematically compared with PI-DeepONet, Vanilla-DeepONet, PINN, and CNN-UNet.  The implementation of SpectONet, and principal findings are summarized as follows:

\begin{enumerate}
    \item[(1)] SpectONet employs deterministic CGL sensor locations to represent input functions using fewer sampling points.

    \item[(2)] SpectONet accurately predicts the displacement responses of three undamped EBB vibration problems under different loading, ICs and BCs.

\item[(3)] SpectONet achieves an average reduction of approximately $79.01\%$ in the mean relative $\mathcal{L}_2$ error compared with the best-performing baseline across the three benchmark problems, with improvements of $94.21\%$, $86.27\%$, and $56.54\%$ for P1, P2, and P3, respectively.

   \item[(4)] The ablation study demonstrates the importance of appropriate loss weighting, collocation-point selection, and combined enforcement of physical constraints.
\end{enumerate}

However, the present study has certain limitations. The numerical investigations are restricted to one-dimensional, linear, undamped EBB vibration problems with deterministic CGL sensor placement. The effects of measurement noise, uncertain material properties, irregular sensor availability, nonlinear structural behavior, and multidimensional geometries have not been investigated. Furthermore, the present analysis does not establish the optimality of the selected sensor locations relative to alternative sampling strategies.
Future research will focus on extending SpectONet to damped and variable-coefficient beam models, Timoshenko beams, plates, shells, and nonlinear structural vibration problems. Additional investigations will explore adaptive sensor placement, noise-robust training, inverse parameter identification, uncertainty quantification, and structural damage detection.

Overall, the numerical findings demonstrate that SpectONet provides an accurate physics-guided operator-learning framework for predicting EBB vibration responses under the considered loading, ICs and BCs.

\section*{Conflict of Interest Statement}
The authors confirm that they have no conflicts of interest related to this paper.

\section*{Funding Declaration}
The authors confirm that they did not receive any funding to carry out this work.

\section*{Authors’ Contributions}\justifying
\textbf{Shivani Saini:} Investigation, Conceptualization, Methodology, Visualization, Software, Experiment,  Writing--original draft \& editing. \textbf{Ramesh Kumar Vats:} Supervision, Formal analysis, Validation,  Writing-review \& editing. \textbf{Arup Kumar Sahoo:} Investigation, Software, Methodology, Validation, Writing-original draft

\section*{Data Availability}
The synthetic generated data will be publicly available at:
\url{https://github.com/SM-a-RT-Lab/SpectONet}.



\section*{Code Availability}

The implementation of the proposed framework will be
publicly available at:
\url{https://github.com/SM-a-RT-Lab/SpectONet}.

\appendix

\section{Appendix}
\label{Apndx}
\addcontentsline{toc}{section}{Appendix}

\setcounter{subsection}{0}
\renewcommand{\thesubsection}{A.\arabic{subsection}}
This appendix presents the training procedures
of the four baseline models: Vanilla-DeepONet,
PI-DeepONet, PINN, and CNN-UNet, as summarized
in Algorithms~\ref{alg:VanillaDeepONet}--\ref{alg:CNNUNet}.
The implementations correspond to the numerical
experiments presented in Section~\ref{sec_5},
while Table~\ref{tab:loss_comparison} compares
the training loss components of the proposed
SpectONet and baseline models.

\begin{table} 
\centering
\caption{Comparison of training objectives used by the proposed method
and baseline models.}
\label{tab:loss_comparison}
\renewcommand{\arraystretch}{1.15}
\setlength{\tabcolsep}{4pt}
\resizebox{\columnwidth}{!}{%
\begin{tabular}{lcccc}
\toprule
\textbf{Model} &
\makecell{\textbf{PDE}\\\textbf{Residual}} &
\makecell{\textbf{BC}\\\textbf{Loss}} &
\makecell{\textbf{IC}\\\textbf{Loss}} &
\makecell{\textbf{Supervised}\\\textbf{Loss}} \\
\midrule
Vanilla-DeepONet
& $\times$ & $\times$ & $\times$ & \checkmark \\
CNN-UNet
& $\times$ & $\times$ & $\times$ & \checkmark \\
PINN
& \checkmark & \checkmark & \checkmark & $\times$ \\
PI-DeepONet
& \checkmark & \checkmark & \checkmark & $\times$ \\
\midrule
\textbf{SpectONet (Ours)}
& \checkmark & \checkmark & \checkmark & $\times$ \\
\bottomrule
\end{tabular}%
}
\end{table}

\subsection{Vanilla-Deep Operator Network}

Algorithm~\ref{alg:VanillaDeepONet} summarizes the
training procedure of the Vanilla-DeepONet~\cite{lu2021learning}
baseline, which relies solely on supervised data
without incorporating the governing PDE, ICs,
or BCs into the loss function. The trained model
is evaluated against the proposed SpectONet
framework. The network architectures and training
hyperparameters for the three benchmark problems
are presented in Table~\ref{Vanilla_Deeponet_parameters}.

\begin{algorithm}
\small
\caption{Training Procedure of Vanilla-DeepONet}
\label{alg:VanillaDeepONet}
\SetAlgoLined

Initialize branch-network parameters $\theta_b$,
trunk-network parameters $\theta_t$, and set
$\theta=\{\theta_b,\theta_t\}$\;

Construct uniformly distributed sensor locations
$\{x_j\}_{j=1}^{N_s}$\;

Generate the training spatial forcing functions
$\{g^{(i)}(x)\}_{i=1}^{N_f}$ and form the corresponding
input sensor vectors
$\mathbf{g}^{(i)}
=[g^{(i)}(x_1),g^{(i)}(x_2),\ldots,
g^{(i)}(x_{N_s})]^T$\;

Generate the training samples
$\{(\mathbf{g}^{(i)},x_j,t_j,u^{(i)}(x_j,t_j))\}$\;

Set convergence tolerance $\varepsilon$\;

\For{each Adam iteration}{

Compute branch features
$\mathbf b=\mathcal B_{\theta_b}(\mathbf g^{(i)})$\;

Compute trunk features
$\mathbf t=\mathcal T_{\theta_t}(x,t)$\;

Predict the solution using Eq.~\eqref{eq:deeponet_operator}\;

Compute the supervised loss using
Eq.~\eqref{eq:deeponet_loss}\;

Update $\theta$ using Adam\;

}

Switch to the L-BFGS optimizer\;

\While{$\mathcal L(\theta)>\varepsilon$}{

Evaluate the supervised loss
$\mathcal L(\theta)$\;

Update $\theta$ using L--BFGS\;

}

\Return{Trained Vanilla-DeepONet model}\;

\end{algorithm}

\begin{table} 
\centering
\caption{Network architecture and hyperparameters employed for the
Vanilla-DeepONet model.}
\label{Vanilla_Deeponet_parameters}

\renewcommand{\arraystretch}{1.15}

\resizebox{\columnwidth}{!}{%
\begin{tabular}{lccc}
\toprule
\textbf{Parameter} & \textbf{P1} & \textbf{P2} & \textbf{P3} \\
\midrule

Activation function
& Tanh & Tanh & Tanh \\

Learning rate
& $1\times10^{-3}$ & $2\times10^{-3}$ & $1\times10^{-2}$ \\

Adam iterations & 500 & 500 & 3000\\
L-BFGS iterations & 5000 & 3000 & 5000\\

Evaluation grid
& $101\times101$ & $101\times101$ & $101\times101$ \\

Branch network architecture
& $[12,10,10]$
& $[12,24,24]$
& $[12,24,24]$ \\

Trunk network architecture
& $[2,10,10]$
& $[2,24,24]$
& $[2,24,24]$ \\

\bottomrule
\end{tabular}%
}

\end{table}

\subsection{Physics-informed Deep Neural Operator}

Algorithm~\ref{alg:PIDeepONet} summarizes the
training procedure of the PI-DeepONet~\cite{cong2026respecting}
baseline, which incorporates the governing PDE,
ICs, and BCs through a physics-informed loss
function. The trained model is evaluated against
the proposed SpectONet framework. The network
architectures and training hyperparameters for
the three benchmark problems are presented in
Table~\ref{PI_Deeponet_parameters}.

\begin{algorithm}
\small
\caption{Training Procedure of PI-DeepONet}
\label{alg:PIDeepONet}
\SetAlgoLined

Initialize branch-network parameters $\theta_b$,
trunk-network parameters $\theta_t$, and set
$\theta=\{\theta_b,\theta_t\}$\;

Construct uniformly distributed sensor locations
$\{x_j\}_{j=1}^{N_s}$\;

Generate the training spatial forcing functions
$\{g^{(i)}(x)\}_{i=1}^{N_f}$ and form the corresponding
input sensor vectors
$\mathbf{g}^{(i)}
=[g^{(i)}(x_1),g^{(i)}(x_2),\ldots,
g^{(i)}(x_{N_s})]^T$\;

Generate interior collocation points $(x_j^r,t_j^r)$,
boundary points $(x_j^b,t_j^b)$,
and initial points $(x_j^0,0)$\;

Set convergence tolerance $\varepsilon$\;

\For{each Adam iteration}{

Compute branch features
$\mathbf b=\mathcal B_{\theta_b}(\mathbf g^{(i)})$\;

Compute trunk features
$\mathbf t=\mathcal T_{\theta_t}(x,t)$\;

Predict the solution using Eq.~\eqref{learned_operator_output}\;

Compute the PDE residual
$R_\theta(g^{(i)})(x,t)$
from Eq.~\eqref{residual_SpectONet} using AD\;

Compute the PDE residual loss $\mathcal L_{PDE}$
using Eq.~\eqref{loss_pde_spectonet}\;

Compute the BC loss $\mathcal L_{BC}$
using Eq.~\eqref{loss_bc_spectonet}\;

Compute the IC loss $\mathcal L_{IC}$
using Eq.~\eqref{loss_ic_spectonet}\;

Compute the total loss using Eq.~\eqref{total_loss}\;

Update $\theta$ using Adam\;

}

Switch to the L-BFGS optimizer\;

\While{$\mathcal L(\theta)>\varepsilon$}{

Evaluate the total loss $\mathcal L(\theta)$\;

Update $\theta$ using L--BFGS\;

}

\Return{Trained PI-DeepONet model}\;

\end{algorithm}

\begin{table} 
\centering
\caption{Network architecture and hyperparameters employed for the PI-DeepONet model.}
\label{PI_Deeponet_parameters}
\renewcommand{\arraystretch}{1.15}
\setlength{\tabcolsep}{4pt}
\resizebox{\columnwidth}{!}{%
\begin{tabular}{lccc}
\toprule
\textbf{Parameter} & \textbf{P1} & \textbf{P2} & \textbf{P3}\\
\midrule
Activation function & Tanh & Tanh & Tanh\\
Learning rate & $1\times10^{-3}$ & $2\times10^{-3}$ & $1\times10^{-2}$\\
Adam iterations & 500 & 500 & 3000\\
L-BFGS iterations & 5000 & 3000 & 5000\\
Evaluation grid & $101\times101$ & $101\times101$ & $101\times101$\\
Branch network architecture
& $[12,10,10]$
& $[12,24,24]$
& $[12,24,24]$\\
Trunk network architecture
& $[2,10,10]$
& $[2,24,24]$
& $[2,24,24]$\\
\bottomrule
\end{tabular}%
}
\end{table}

\subsection{Physics-informed Neural Network}

Algorithm~\ref{alg:PINN} summarizes the training
procedure of the PINN~\cite{raissi2019physics}
baseline, which incorporates the governing PDE,
ICs, and BCs into a physics-informed loss function
without employing an operator learning architecture.
The network architectures and training hyperparameters
for the three benchmark problems are presented in
Table~\ref{PINN_parameters}.

\begin{algorithm} 
\small
\caption{Training Procedure of PINN}
\label{alg:PINN}
Generate interior collocation points
$(x_i^r,t_i^r)$, boundary points
$(x_i^b,t_i^b)$, and initial points
$(x_i^0,0)$\;

Set convergence tolerance $\varepsilon$\;

Train the network using the Adam optimizer\;

\For{each Adam iteration}{
Evaluate the governing PDE residual from Eq.~\eqref{eq:pinn_residual}
using AD\;

Compute the residual loss using Eq.~\eqref{eq:pinn_pde_loss}\;

Compute the IC loss using Eq.~\eqref{eq:pinn_ic_loss}\;

Compute the BC loss using Eq.~\eqref{eq:pinn_bc_loss}\;

Compute the composite physics-informed loss Eq.~\eqref{total_loss}\;

Update the network parameters
$\theta$ using Adam\;

}

Switch to the L-BFGS optimizer\;

\While{$\mathcal L(\theta)>\varepsilon$}{

Evaluate the composite loss
$\mathcal L(\theta)$\;

Update $\theta$ using L--BFGS\;

}

\Return{Trained PINN model}\;

\end{algorithm}

\begin{table} 
\centering
\caption{Network architecture and hyperparameters employed for the PINN model.}
\label{PINN_parameters}
\renewcommand{\arraystretch}{1.5}
\setlength{\tabcolsep}{4pt}
\resizebox{\columnwidth}{!}{%
\begin{tabular}{lccc}
\toprule
\textbf{Parameter} & \textbf{P1} & \textbf{P2} & \textbf{P3}\\
\midrule
Activation function & Tanh & Tanh & Tanh\\
Learning rate & $1\times10^{-3}$ & $2\times10^{-3}$ & $1\times10^{-2}$\\
Adam iterations & 500 & 500 & 3000\\
L-BFGS iterations & 5000 & 3000 & 5000\\
Evaluation grid & $101\times101$ & $101\times101$ & $101\times101$\\
Network architecture
& $[2,10,10,10,1]$
& $[2,37,37,1]$
& $[2,21,21,21,21,21,1]$\\
\bottomrule
\end{tabular}%
}
\end{table}

\subsection{U-Net Convolutional Neural Network}

Algorithm~\ref{alg:CNNUNet} summarizes the training
procedure of the CNN-UNet~\cite{ronneberger2015u}
baseline. The model receives the spatio-temporal
grid as input and learns the corresponding
displacement field through supervised training
using the reference solution. The network
architectures and training hyperparameters
for the three benchmark problems are summarized
in Table~\ref{CNN_UNet_parameters}.

\begin{algorithm}
\small
\caption{Training Procedure of CNN-UNet}
\label{alg:CNNUNet}
\SetAlgoLined

Initialize the encoder and decoder parameters $\theta$\;

Construct the space-time input grid
$\mathbf{X}=(x,t)$\;

Generate the corresponding reference solution field
$\mathbf{u}$\;

Set the convergence tolerance $\varepsilon$\;

\For{each Adam iteration}{

Feed the input grid $\mathbf{X}$ into the encoder\;

Extract hierarchical latent feature maps at different
resolution levels\;

Pass the latent feature maps and skip connections
through the decoder\;

Reconstruct the displacement field
$\hat{u}_{\theta}(x,t)$\;

Compute the supervised MSE loss
$\mathcal{L}(\theta)
=\dfrac{1}{N_q}
\displaystyle\sum_{j=1}^{N_q}
\left|
\hat{u}_{\theta}(x_j,t_j)
-u(x_j,t_j)
\right|^2$\;

Update the network parameters $\theta$ using Adam\;

}

Switch to the L-BFGS optimizer\;

\While{$\mathcal{L}(\theta)>\varepsilon$}{

Evaluate the supervised training loss
$\mathcal{L}(\theta)$\;

Update the network parameters $\theta$
using L--BFGS\;

}

\Return{Trained CNN-UNet model}\;

\end{algorithm}


\begin{table} 
\centering
\caption{Network architecture and hyperparameters employed for the CNN-UNet model.}
\label{CNN_UNet_parameters}
\renewcommand{\arraystretch}{1.15}
\setlength{\tabcolsep}{4pt}
\resizebox{\columnwidth}{!}{%
\begin{tabular}{lccc}
\toprule
\textbf{Parameter} & \textbf{P1} & \textbf{P2} & \textbf{P3}\\
\midrule
Activation function & Tanh & Tanh & Tanh\\
Learning rate & $1\times10^{-3}$ & $2\times10^{-3}$ & $1\times10^{-2}$\\
Adam iterations & 500 & 500 & 3000\\
L-BFGS iterations & 5000 & 3000 & 5000\\
Evaluation grid & $101\times101$ & $101\times101$ & $101\times101$\\
Input channels & 2 & 2 & 2\\
Encoder-1 channels & 1 & 4 & 4\\
Encoder-2 channels & 2 & 8 & 8\\
Bottleneck channels & 2 & 8 & 8\\
Decoder input channels & 3 & 12 & 12\\
Decoder output channels & 1 & 4 & 4\\
Output channels & 1 & 1 & 1\\
Kernel size & $3\times3$ & $3\times3$ & $3\times3$\\
\bottomrule
\end{tabular}%
}
\end{table}


\bibliographystyle{elsarticle-num} 
\bibliography{Refrences.bib} 

\end{document}